%% file: main.tex
\documentclass[10pt,twocolumn,letterpaper]{article}

\usepackage{cvpr}              % To produce the CAMERA-READY version
\usepackage{multirow}
\usepackage{ulem}
\usepackage{algpseudocode}
\usepackage{algorithm}
\usepackage{wrapfig}
\usepackage{comment}
\usepackage{pifont}
\usepackage{microtype}
\usepackage{boldline}
\usepackage[table]{xcolor}
\usepackage{longtable}
\usepackage{arydshln}

\usepackage{cuted}
\usepackage{afterpage}

\usepackage{pifont}% http://ctan.org/pkg/pifont
\newcommand{\cmark}{\ding{51}}%
\newcommand{\xmark}{\ding{55}}%

\usepackage{marvosym}

\definecolor{cvprblue}{rgb}{0.21,0.49,0.74}
\usepackage[pagebackref,breaklinks,colorlinks,allcolors=cvprblue]{hyperref}

\definecolor{headergray}{gray}{0.9}

\def\paperID{******} % *** Enter the Paper ID here
\def\confName{CVPR}
\def\confYear{2026}

\newcommand{\nickname}{PhysInOne}

\renewcommand{\etal}{\textit{et al}.}
\renewcommand{\ie}{\textit{i}.\textit{e}.}
\renewcommand{\eg}{\textit{e}.\textit{g}.}
\renewcommand{\etc}{\textit{etc}.}

\title{\nickname{}: Visual Physics Learning and Reasoning in One Suite}

\begin{document}

\author{
Siyuan Zhou$^{1}$\textsuperscript{\dag}\phantom{x}
Hejun Wang$^1$\textsuperscript{\dag}\phantom{x}
Hu Cheng$^1$\textsuperscript{\dag}\phantom{x}
Jinxi Li$^1$\textsuperscript{\dag}\phantom{x}
Dongsheng Wang$^1$\textsuperscript{\dag}\phantom{x}
Junwei Jiang$^1$\textsuperscript{\dag}\phantom{x}\\
Yixiao Jin$^1$\textsuperscript{\dag}\phantom{x}
Jiayue Huang$^1$\textsuperscript{\dag}\phantom{x}
Shiwei Mao$^1$\textsuperscript{\dag}\phantom{x}
Shangjia Liu$^2$\phantom{x}
Yafei Yang$^1$\phantom{x} \\
Hongkang Song$^1$\phantom{x}
Shenxing Wei$^1$\phantom{x}
Zihui Zhang$^1$\phantom{x}
DataTeam$^1$\textsuperscript{*}\phantom{x} \\
Bing Wang$^2$\phantom{x}
Zhihua Wang$^3$\phantom{x}
Chuhang Zou$^4$\textsuperscript{\Letter}\phantom{x}
Bo Yang$^1$\textsuperscript{\Letter} \\ \\
 \textsuperscript{1} vLAR Group,  \textsuperscript{2} The Hong Kong Polytechnic University\quad  \textsuperscript{3} Syai Singapore \quad  \textsuperscript{4} Meta \\
\textsuperscript{\dag} equal contribution and co-first authorship \quad \textsuperscript{\Letter} corresponding authors \\ 
%\begin{center}
\makebox[0pt][c]{
\textsuperscript{*}{\small
\begin{minipage}{1.005\textwidth}\centering
\{Peng Huang,
Shijie Liu,
Zhengli Hao,
Hao Li, 
Yitian Li,
Wenqi Zhou,
Zhihan Zhao,
Zongqi He, 
Hongtao Wen,
Shouwang Huang,
Peng Yun\\
Bowen Cheng,
Pok Kazaf Fu,
Wai Kit Lai,
Jiahao Chen,
Kaiyuan Wang,
Zhixuan Sun,
Ziqi Li,
Haochen Hu,
Di Zhang,
Chun Ho Yuen\}
\end{minipage}
}
%\end{center}
}\\
{\tt\small \{siyuan.zhou, hejun.wang, hu123.cheng, jinxi.li\}@connect.polyu.hk, bo.yang@polyu.edu.hk}\\
{\mbox{\url{https://vlar-group.github.io/PhysInOne.html}} }
}

\twocolumn[{%
\renewcommand\twocolumn[1][]{#1}%
    \maketitle
    \begin{center}
        \vspace{-15pt}
        \centering
        \includegraphics[width=1.0\linewidth]{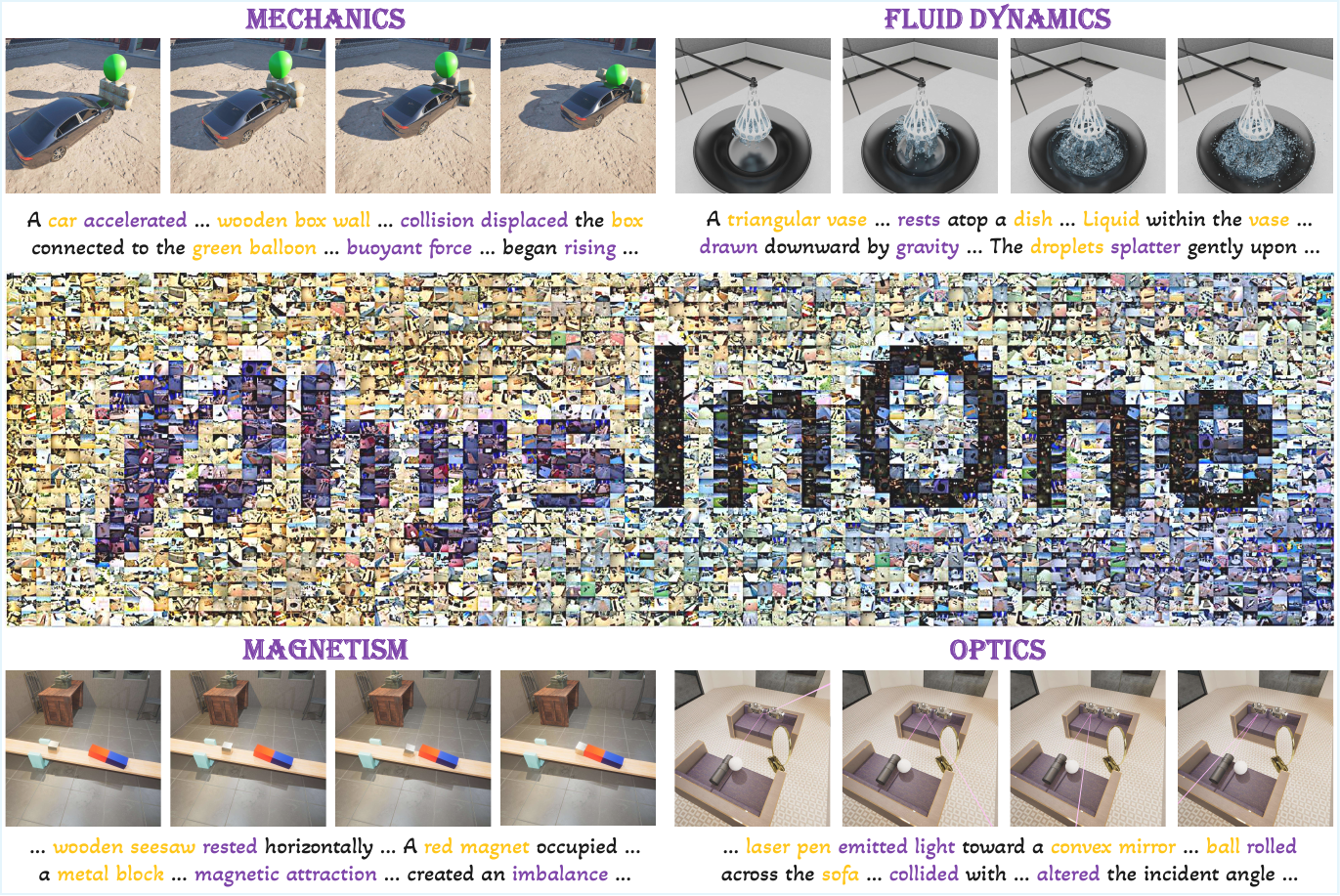}
        \vspace{-20pt}
        \captionof{figure}{We present \nickname{}, a large-scale dataset of 153,810 dynamic 3D scenes and 2 million annotated videos, systematically capturing 71 basic physical phenomena and multiobject interactions to advance visual physics learning and reasoning.
        }
        \label{fig:opening}
        \vspace{0pt}
    \end{center}
}]

% \maketitle
% \vspace{-0.5cm}
\begin{abstract}

\vspace{-7.9pt}
We present \textbf{\nickname{}}, a large-scale synthetic dataset addressing the critical scarcity of physically-grounded training data for AI systems. Unlike existing datasets limited to merely hundreds or thousands of examples, \nickname{} provides 2 million videos across 153,810 dynamic 3D scenes, covering 71 basic physical phenomena in mechanics, optics, fluid dynamics, and magnetism. Distinct from previous works, our scenes feature multiobject interactions against complex backgrounds, with comprehensive ground-truth annotations including 3D geometry, semantics, dynamic motion, physical properties, and text descriptions. We demonstrate \nickname{}’s efficacy across four emerging applications: physics-aware video generation, long-/short-term future frame prediction, physical property estimation, and motion transfer. Experiments show that fine-tuning foundation models on \nickname{} significantly enhances physical plausibility, while also exposing critical gaps in modeling complex physical dynamics and estimating intrinsic properties. As the largest dataset of its kind, orders of magnitude beyond prior works, \nickname{} establishes a new benchmark for advancing physics-grounded world models in generation, simulation, and embodied AI. 
\vspace{-12pt}
\end{abstract}
% \pagebreak[4]

\input{sec/1_intro}
\input{sec/2_liter}

\input{sec/3_dataset}

\input{sec/4_applications}

\vspace{-0.1cm}
\input{sec/5_conclusion}

\clearpage
\section*{Acknowledgments}
This work was supported in part by Research Grants Council of Hong Kong under Grants 15219125 \& 15225522. \\

\noindent We would like to express our sincere gratitude to (in alphabetical order) Geer Chen, Jinhe Chen, Zhiyuan Chen, Yuanhaonan Deng, Shuo Feng, Wenxuan Guo, Junpeng Hu, Ruitao Hu, Ying Ji, Yixuan Jiang, Jiani Liu, Xinjie Liu, Xinsheng Liu, Jiyuan Ma, Qiyue Ma, Chenyang Mao, Yukun Miao, Ye Peng, Yuanyue Qiao, Dacheng Qin, Xiangnuo Ren, Xiaowen Song, Jingqi Tian, Hong Wang, Huixuechun Wang, Zheng Wang, Weipeng Wu, Zhaowei Wu, Kai Xing, Ran Yan, Leize Yang, Ruizhe Yang, Ao Yu, and Minhao Zhu for their essential contributions and dedicated efforts in conducting human evaluations for \nickname{} benchmark.

{
\small
\bibliographystyle{ieeenat_fullname}
\bibliography{references}
}

\input{sec/X_suppl}

\end{document}

%% file: sec/1_intro.tex
\section{Introduction}
\label{sec:intro}

Large-scale, high-quality data serves as the catalyst for modern machine learning breakthroughs. In natural language processing, massive corpora like Common Crawl \cite{CommonCrawl} have enabled the unprecedented abilities of LLMs such as GPT-4 \cite{OpenAI2023}. A similar paradigm has driven 2D computer vision: foundational datasets like ImageNet \cite{Deng2009} and MS-COCO \cite{Lin2014} accelerated discriminative tasks, while massive-scale collections like LAION-5B \cite{Schuhmann2022} have revolutionized image generation. Extending this to the 3D/4D domain, vast repositories such as Objaverse-XL \cite{Deitke2023,Deitke2023a} have facilitated rapid advances in reconstruction, while Panda-70M \cite{Chen2024} has become a cornerstone for video generation and world models.

Despite impressive results demonstrated by state-of-the-art LLM/LVM/MLLM/GenAI models trained on these datasets, as shown in \cite{Motamed2025,Kang2024}, their ability to truly comprehend the physical world and its laws remains limited, a challenge that has long been regarded as a major milestone on the path toward general-purpose AI. For example, while visually appealing, videos generated by AI frequently violate basic physical laws, with objects falling upward or suddenly shifting velocity. This has spurred growing interest in learning physics from visual and/or texts in recent works \cite{Li2023c,Li2023,Furuta2024,Cai2024,Kaneko2024,Zhu2024,Liu2024a,Li2025c,Yuan2025,Zhan2025,Balazadeh2025,Garcia2025,Zhao2025,Li2025b,Lin2025}. Despite achieving encouraging results, these models are typically trained on only dozens, hundreds, or thousands of data points, which restricts them to limited physical phenomena, making them far from true world simulators. This limitation primarily stems from the lack of large-scale, high-quality training data that capture various physical objects and scenes, and encompass diverse physical phenomena in 3D space.

In this paper, we present \textbf{\nickname{}}, a large-scale synthetic dataset featuring richly annotated dynamic 3D scenes. These scenes depict a wide range of physical phenomena among multiple objects interacting against complex backgrounds. Drawing on the introductory university-level textbook \textit{Fundamentals of Physics} \cite{Halliday2021} and recent research \cite{Motamed2025,Meng2025}, we focus on visual learning in four core areas of everyday physics: mechanics, optics, fluid dynamics, and magnetism, while omitting thermodynamics and acoustics,
which are rarely visible or require additional sensory data like temperature or sound beyond the scope of our dataset.

Across these four areas, we identify 71 key physical phenomena commonly encountered in daily life, such as gravity, reflection, buoyancy, magnetic attraction, \textit{etc.}, covering the vast majority, if not all, of relevant everyday physics. For these phenomena, we create 153,810 dynamic 3D scenes. Each scene consists of multiple objects and undergoes multiple physical phenomena simultaneously or sequentially, all strictly governed by fundamental physical laws including Newton's Laws, Mass Conservation, (Angular) Momentum Conservation, Hooke's Law, \etc{}. For each dynamic 3D scene, we record 13 videos capturing the entire physical activity: 12 from fixed cameras and 1 from a smoothly moving monocular camera. Alongside these videos, we manually annotate a high-quality paragraph of text, accurately describing the scene's visual elements and activity. This yields a comprehensive dataset of 2 million dynamic videos with ground truth annotations, including object 3D meshes and moving trajectories, 2D masks, materials, depths, text descriptions, camera poses, \etc{}, being orders of magnitude larger than all existing visual physics datasets. 

With this significant increase in scale and diversity, we believe \nickname{} has the potential to make a substantial impact on research progress in visual physics learning. In this paper, we present four emerging application examples by exploiting our dataset, including: 

\textbf{1) Physics-aware Video Generation}: Video generation is often touted as a promising path towards simulating the physical world \cite{OpenAI2024}, but currently still lacks the essential ability to predict physically plausible clips. By fine-tuning SVD \cite{Blattmann2023}, CogVideoX \cite{Yang2025}, and WAN \cite{Wan2025} models using just a moderate subset of text-video pairs from \nickname{}, we significantly boost the physical plausibility of generated videos. This improvement is evidenced by our newly introduced metric focused on physical motion fidelity.   

\textbf{2) (Long-/Short-term) Future Frame Prediction}: As physicist Richard Feynman stated, ``What I cannot create, I do not understand". This insight implies that predicting how videos continue is a crucial task to test a model's understanding of physical principles. This ability also serves as a key enabler for emerging applications in future-aware robot planning and embodied AI \cite{Ding2025}. To this end, we showcase this critical task on \nickname{} by training and evaluating a series of recent models, including TiNeuVox \cite{Fang2022}, DefGS \cite{Yang2024c}, FreeGave \cite{Li2025c}, TRACE \cite{Li2025b}, ExtDM \cite{Zhang2024}, and MAGI-1 \cite{AI2025}.

\begin{table*}[t]\tabcolsep=0.15cm
\centering
\caption{A comparison between \nickname{} and existing datasets that are relevant to learning physics from dynamic videos. \nickname{} provides massive 3D scenes with complex objects and backgrounds across diverse physical phenomena, surpassing all prior works.  }\vspace{-0.2cm}
\resizebox{1\linewidth}{!}{
\renewcommand{\arraystretch}{1.15}
\begin{tabular}{r|r|c|c|c|c|c|c|c|c|c} \hline
    & \multirow{2}{*}{\begin{tabular}[c]{@{}c@{}} \textbf{Year} \end{tabular}}
    & \multirow{2}{*}{\begin{tabular}[c]{@{}c@{}} \textbf{\# Physical} \\ \textbf{Phenomena}\end{tabular}}
    & \multirow{2}{*}{\begin{tabular}[c]{@{}c@{}} \textbf{\# Physical} \\ \textbf{Scenes}\end{tabular}}
    & \multirow{2}{*}{\begin{tabular}[c]{@{}c@{}} \textbf{\# Dynamic} \\ \textbf{Videos}\end{tabular}}
    &\multirow{2}{*}{\begin{tabular}[c]{@{}c@{}} \textcolor{orange}{\textbf{Multiobject}} \\ \textcolor{orange}{\textbf{Interactions}} \end{tabular}}
    &\multirow{2}{*}{\begin{tabular}[c]{@{}c@{}} \textcolor{violet}{\textbf{Multiphysics}} \\ \textcolor{violet}{\textbf{Activities}} \end{tabular}}
    &\multirow{2}{*}{\begin{tabular}[c]{@{}c@{}} \textcolor{magenta}{\textbf{Complex}} \\ \textcolor{magenta}{\textbf{Objects}} \end{tabular}}
    &\multirow{2}{*}{\begin{tabular}[c]{@{}c@{}} \textcolor{blue}{\textbf{Complex}} \\ \textcolor{blue}{\textbf{Backgrounds}} \end{tabular}}
    & \multirow{2}{*}{\begin{tabular}[c]{@{}c@{}} \textbf{\# Fixed / Monocular} \\ \textbf{Cameras} \end{tabular}}
    & \multirow{2}{*}{\begin{tabular}[c]{@{}c@{}} \textbf{Annotations} (\textbf{G}eometry, \textbf{S}emantics, \\ \textbf{M}otion, \textbf{P}roperties, \textbf{T}exts) \end{tabular}}  \\
   & & & & &  & & & &\\ \hline 
    
Physics101 \cite{Wu2016} & BMVC'16 & 4 & - & 17,408 & \textcolor{red}{\xmark} & \textcolor{red}{\xmark} & \textcolor{red}{\xmark} & \textcolor{red}{\xmark} & 3, 0 & P \\ \hline

ShapeStacks \cite{Groth2018}& ECCV'18 & 1 & 20,000 & 20,000 & \textcolor{green}{\cmark} & \textcolor{red}{\xmark} & \textcolor{red}{\xmark} & \textcolor{red}{\xmark} & 1, 0 & S, M, T \\ \hline

ScalarFlow \cite{Eckert2019}& TOG'19 & 1 & 100 & 100 & \textcolor{red}{\xmark} & \textcolor{red}{\xmark} & \textcolor{red}{\xmark} & \textcolor{red}{\xmark} & - , - & M, P \\ \hline

CLEVRER \cite{Yi2020}& ICLR'20 & 1 & 10,000 & 10,000 & \textcolor{green}{\cmark} & \textcolor{red}{\xmark} & \textcolor{red}{\xmark} & \textcolor{red}{\xmark} & - , - & M, T \\ \hline

Physion \cite{Bear2021}& NeurIPS'21 & 8 & - & 24,000 & \textcolor{green}{\cmark} & \textcolor{red}{\xmark} & \textcolor{red}{\xmark} & \textcolor{green}{\cmark} & - , - & M, T \\ \hline

ComPhy \cite{Chen2022}& ICLR'22 & 1 & 12,000 & 12,000 & \textcolor{green}{\cmark} & \textcolor{red}{\xmark} & \textcolor{red}{\xmark} & \textcolor{red}{\xmark} & 1, 0 & M, T \\ \hline

SPACE+ \cite{Duan2022}& ECCV'22 & 3 & 15,000 & 57,057 & \textcolor{green}{\cmark} & \textcolor{red}{\xmark} & \textcolor{red}{\xmark} & \textcolor{red}{\xmark} & 1, 0 & G, S, M \\ \hline

Physion++ \cite{Tung2023}& NeurIPS'23 & 9 & - & $\sim$ 18,000 & \textcolor{green}{\cmark} & \textcolor{red}{\xmark}  & \textcolor{red}{\xmark} & \textcolor{red}{\xmark} & - , -  & P, T \\ \hline
PAC-NeRF \cite{Li2023} & NeurIPS'23 & - & 9 & 99 & \textcolor{red}{\xmark} & \textcolor{red}{\xmark} & \textcolor{red}{\xmark} & \textcolor{red}{\xmark} & 11, 0 & P \\ \hline

VideoPhy \cite{Bansal2025}& ICLR'25 & - & - & 11,330 & \textcolor{green}{\cmark} & - & \textcolor{green}{\cmark} & \textcolor{green}{\cmark} & 0, 1 & T\\ \hline

PhysTwin \cite{Jiang2025}& ICCV'25 & - & 22 & 66 & \textcolor{green}{\cmark} & \textcolor{red}{\xmark} & \textcolor{green}{\cmark} & \textcolor{red}{\xmark} & 3, 0 & M, P\\ \hline

IntPhys2 \cite{Bordes2025}& arXiv'25 & - & 344 & 1416 & \textcolor{green}{\cmark} & \textcolor{red}{\xmark} & \textcolor{green}{\cmark} & \textcolor{green}{\cmark} & - , -& T \\ \hline 

Physics-IQ \cite{Motamed2025}& arXiv'25 & - & 66 & 396 & \textcolor{green}{\cmark} & \textcolor{red}{\xmark} & \textcolor{green}{\cmark} & \textcolor{red}{\xmark} & 3, 0 & M \\ \hline

PhysVid \cite{Zhan2025}& arXiv'25 & 3 & - & $\sim$33,000 & \textcolor{red}{\xmark}  & \textcolor{red}{\xmark}  & \textcolor{red}{\xmark} & \textcolor{red}{\xmark} & - , -& S, M, P\\ [+0.1em] \hline 

NewtonGen \cite{Yuan2025}& ICLR'26 & 12 & - & 1200 & \textcolor{red}{\xmark}  & \textcolor{red}{\xmark}  & \textcolor{green}{\cmark} & \textcolor{green}{\cmark} & 1 , -& T \\ [+0.1em] \hlineB{3}

\textbf{\nickname{} (Ours)} & CVPR'26 & \textbf{71} & \textbf{153,810} & \textbf{2 million} & \textcolor{green}{\cmark} & \textcolor{green}{\cmark} & \textcolor{green}{\cmark} & \textcolor{green}{\cmark} & 12, 1 & G, S, M, P, T \\ [+0.1em] 
\bottomrule
\end{tabular}
}
\label{tab:existing_datasets}
\vspace{-0.3cm}
\end{table*}

\textbf{3) Physical Properties Estimation}: Inferring physical properties from visual observations, a.k.a system identification or inverse physics, aims to learn disentangled, interpretable, and editable physical quantities, thus enabling various applications like resimulation, scene editing, robot manipulation, \etc{}. We benchmark this task on our dataset using two recent representative methods PAC-NeRF \cite{Li2023} and GIC \cite{Cai2024} to estimate a variety of parameters like Young's modulus, Poisson's ratio, viscosity, \etc{}.  

\textbf{4) Motion Transfer}: Transferring motions between scenes enables entertaining image animations and controllable generation. By evaluating recent methods MotionPro \cite{Zhang2025} and GoWithTheFlow \cite{Burgert2025} on \nickname{}, we showcase the pipeline of transferring physically meaningful motions from one scene to another. However, the results remain largely unsatisfactory, underscoring the unique challenge in transferring complex physical motion patterns.

Overall, these findings, enabled by our large-scale, high-quality \nickname{} dataset, represent just a fraction of its potential. We anticipate the research community will leverage it to accelerate progress in 3D/4D vision and beyond. 

%% file: sec/2_liter.tex
\section{Related Datasets and Benchmarks}
\label{sec:liter}

\phantom{xw}\textbf{Datasets for Learning Physical Dynamics}: Existing datasets for learning and reasoning physical dynamics often focus on a narrow set of physical phenomena. For instance, ShapeStacks \cite{Groth2018} and StableText2Brick \cite{Pun2025} datasets only address object or part stability, while datasets like
VOE \cite{Piloto2018},
IntPhys2019 \cite{Riochet2022}, 
PHYRE \cite{Bakhtin2019},
ADEPT \cite{Smith2019}, 
CLEVRER \cite{Yi2020},
ESPRIT \cite{Rajani2020},
Physion \cite{Bear2021},
CoPhy \cite{Baradel2022},
CRAFT \cite{Ates2022},
and SPACE+ \cite{Duan2022}
examine simple scenarios where single object trajectory follows intuitive physics such as object persistence/continuity/solidity, rather than rigorously verifying adherence to a wide range of strict physical laws. In addition, almost all objects in these datasets are oversimplified shapes (cubes, balls, \etc{}) with homogeneous colors against clean backgrounds. Other datasets specialize in learning fluid dynamics from visual data: ScalarFlow \cite{Eckert2019}, TomoFluid \cite{Zang2020}, and FLAME \cite{Shamsoshoara2021} focus on smoke, while Obstacles \cite{Baieri2023},
Richter \etal{} \cite{Richter2022}, 
and NeuroFluid \cite{Guan2022}
on liquids. 

\textbf{Datasets for Learning Physical Properties}: A growing number of datasets have been collected for estimating different physical properties \cite{Wang2018a,Wang2018l,Wu2016}, including Physics101 \cite{Wu2016} for mass and volume, 
Physion++ \cite{Tung2023},
Materialistic \cite{Sharma2023},
PAC-NeRF \cite{Li2023},
PhysTwin \cite{Jiang2025},
PhysXNet \cite{Cao2025},
SOPHY \cite{Cao2025a},
PixieVerse \cite{Le2025},
PhysVid \cite{Zhan2025},
and VoMP \cite{Dagli2025},
for various physical properties like mass, friction, elasticity, deformability, \etc{}. These datasets usually focus on isolated individual objects or oversimplified 3D scenes.

\textbf{Datasets for Testing Physical Understanding}: Recent advances of large language, vision, multimodal, video, and world models have led to a series of benchmarking datasets to test the physical understanding abilities of these models, including
ComPhy \cite{Chen2022},
PerceptionTest \cite{Patraucean2023},
TraySim \cite{Cherian2024},
GRASP \cite{Jassim2024}, 
PhyBench \cite{Meng2024},
Physics-IQ \cite{Motamed2025},
Morpheus \cite{Zhang2025a},
WorldModelBench \cite{Li2025a},
WISA \cite{Wang2025},
VBench-2.0 \cite{Zheng2025},
PhysBench \cite{Chow2025},
DynSuperCLEVR \cite{Wang2025a},
VideoPhy \cite{Bansal2025},
VideoPhy-2 \cite{Bansal2025a},
PhyX \cite{Shen2025},
IntPhys2 \cite{Bordes2025},
PhyGenBench \cite{Meng2025},
UGPhysics \cite{Xu2025},
STI-Bench \cite{Li2025}, 
PhyWorldBench \cite{Gu2025},
PisaBench \cite{Li2025e},
NewtonGen \cite{Yuan2025},
and NewtonBench-60K \cite{Le2025a}.
These benchmarks typically focus on text–image–video question answering (QA) or visual understanding and generation in narrow scenarios, and thus lack the precise and diverse visual data and annotations needed to train or fine-tune large, general models for quantitatively evaluating visual physics learning.

Table \ref{tab:existing_datasets} compares \nickname{} with key existing datasets designed for physics learning from videos, highlighting our scale and diversity across multiple dimensions.

%% file: sec/3_dataset.tex
\begin{figure*}[t!]
\centering
\includegraphics[width=1.\linewidth]{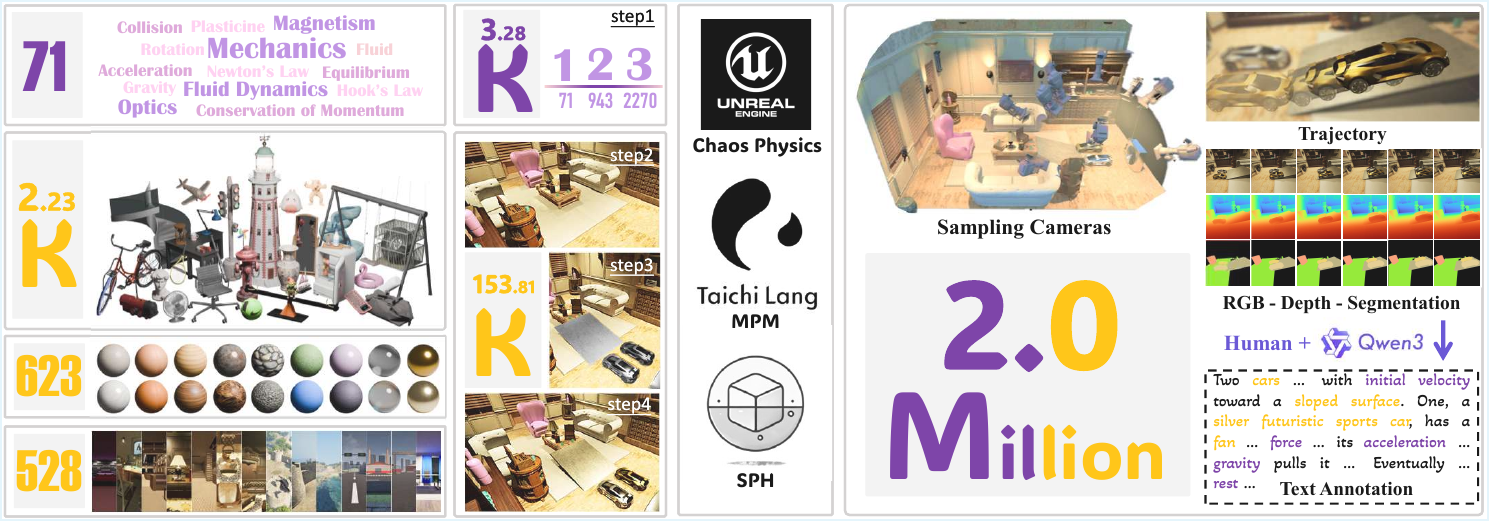}
\vskip -0.1in
\caption{Spanning 71 basic physical phenomena scaled to 3,284 multiphysics activities, \nickname{} comprises 153,810 unique scenes featuring 2,231 objects across 528 backgrounds and 623 materials, yielding 2 million videos with comprehensive annotations.}
\label{fig:dataset_pipeline}
\vspace{-0.4cm}
%\vskip 0.4in
\end{figure*}

\section{\nickname{}}
\label{sec:dataset}
The pipeline to construct our dataset involves multiple stages as detailed in the following sections and shown in Figure \ref{fig:dataset_pipeline}. 

\subsection{Physical Phenomena and Laws}\label{sec:dataset_physics}

Within the four core areas of everyday physics, mechanics, optics, fluid dynamics, and magnetism, we identify 71 basic physical phenomena, covering the predominant physics commonly observed in daily life. Each basic phenomenon is governed by one or more fundamental physical laws, such as Newton's Laws, Conservation of Momentum, Hooke's Law, Bernoulli's Principle, \etc{}. A complete list of these basic phenomena and their governing laws is in Appendix \ref{app:phys_phen_laws}.

Importantly, each basic physical phenomenon represents a conceptual occurrence in daily life, not a specific, concrete 3D scene. In Section \ref{sec:dataset_scenes}, we will create tangible 3D scenes that instantiate one or more of these basic phenomena.

\subsection{Collecting 3D Assets}\label{sec:dataset_assets}

In real-world scenarios, a physical 3D scene typically comprises multiple objects interacting with each other against complex backgrounds. The dynamics of such scenes instantiate one or more basic physical phenomena and are strictly governed by the fundamental physical laws identified in Section \ref{sec:dataset_physics}. To replicate as many diverse examples of these complex 3D scenes as possible, we collect a rich set of 3D assets for use in Section \ref{sec:dataset_scenes}. These assets include:

\begin{itemize}[leftmargin=*] %\vspace{-0.2cm}
\setlength{\itemsep}{1pt}
\setlength{\parsep}{1pt}
\setlength{\parskip}{1pt}
    \item \textbf{3D Objects}: 
    Since our dataset targets everyday physics and particularly the identified 71 basic physical phenomena, we collect 2,231 common objects ($\sim$163 categories) from Sketchfab \cite{Sketchfab}, FAB \cite{FAB}, and BlenderKit \cite{BlenderKit}. These objects, directly relevant to physical interactions, distinguish our dataset from those like Objaverse/XL designed for generic 3D tasks. The collection includes:
    \begin{itemize}[leftmargin=*] %\vspace{-0.2cm}
    \setlength{\itemsep}{1pt}
    \setlength{\parsep}{1pt}
    \setlength{\parskip}{1pt}
        \item \textit{solid objects}: like hammers and bricks that maintain their shape during physical interactions.
        \item \textit{interactable objects}: which have  movable parts (\eg{}, rotatable fans), implemented as Blueprints in UE \cite{UnrealEngine}. 
        \item \textit{destructible objects}: such as breakable glasses, implemented as Geometry Collections in UE.
        \item \textit{deformable objects}: whose shape changes during physical activities. Their deformation is simulated using MPM \cite{Hu2018}, as detailed in Section \ref{sec:dataset_dynamics}. 
        \item \textit{granular objects}: primarily representing granular materials like sand, also simulated using MPM. 
        \item \textit{liquid}: like water, cream, milk, \etc{}, simulated using SPH \cite{Gingold1977}, as detailed in Section \ref{sec:dataset_dynamics}. 
    \end{itemize}
    \item \textbf{Materials}: To achieve greater realism and diversity in object representation, we further collect 623 materials across 5 categories: plastic, metal, wood, stone, and fabric. Each material’s adjustable physical properties, such as friction coefficients, density, restitution, enables the creation of numerous scene variations, as discussed in Section \ref{sec:dataset_scenes}.
    \item \textbf{3D Backgrounds}: To provide realistic environments where diverse physical phenomena occur, we collect 528 distinct 3D backgrounds. These encompass various indoor and outdoor scenarios at different spatial scales, such as living rooms, bedrooms, factories, swimming pools, \etc{}.   
\end{itemize} %\vspace{-0.2cm}
Notably, all collected 3D assets are freely available and licensed for both academic and commercial use. More details of 3D assets are provided in Appendix \ref{app:3d_assets}.

\subsection{Creating Multiphysics Multiobject 3D Scenes}\label{sec:dataset_scenes}
With the collected 3D assets and identified 71 basic physical phenomena, we create massive, realistic, and challenging 3D scenes through the following steps:
\begin{itemize}[leftmargin=*] %\vspace{-0.2cm}
\setlength{\itemsep}{1pt}
\setlength{\parsep}{1pt}
\setlength{\parskip}{1pt}
    \item \textit{Step 1: Emulating Multiphysics}: Since daily physical activities often involve one or more basic physical phenomena simultaneously or sequentially, we emulate 3 categories:
    \begin{itemize}[leftmargin=*] %\vspace{-0.2cm}
    \setlength{\itemsep}{1pt}
    \setlength{\parsep}{1pt}
    \setlength{\parskip}{1pt}
        \item \textit{single-physics activities}: We treat each of the 71 basic phenomena as a distinct single-physics activity. 
        \item \textit{double-physics activities}: We combine 2 basic phenomena into a complex activity. While $C_{71}^2$ combinations exist, many lack physical meaning. After careful selection, we create 943 double-physics activities.
        \item \textit{triple-physics activities}: Similarly, we combine 3 basic phenomena into a more complex activity. After selection, we create 2270 valid triple-physics activities. 
    \end{itemize}
    The resulting 71 + 943 + 2270 = 3284 physical activities are still conceptual at this stage; they are implemented as tangible 3D scenes in subsequent steps.  
    \item \textit{Step 2: Setting Up Backgrounds}: We set up physically plausible backgrounds aligned with the target activity. 
    \item \textit{Step 3: Placing Multiobjects}: For each activity, we create many tangible 3D scenes. In each scene, we select one or more 3D assets and place them into a background.  
    \item \textit{Step 4: Varying Materials}: For 3D assets in each scene, we alter their materials and vary physical properties.  
\end{itemize}

Following the above four steps, we create 153,810 unique multiphysics multiobject 3D scenes. Each activity is realized through an average of 46.84 distinct scenes featuring varied assets, materials, and backgrounds. The average number of 3D assets (objects) per scene increases with activity complexity: 3.9/6.3/7.8 for single-/double-/triple- physics activities, demonstrating the escalating scene complexity. More details are in Appendix \ref{app:3d_scenes}. 

\subsection{Simulating Physical Dynamics}\label{sec:dataset_dynamics}

To simulate physical activities of our 153,810 scenes as accurate as possible, we employ multiple simulation algorithms, packages and engines. In particular, \textit{Chaos Physics} integrated in UE5 \cite{UnrealEngine} handles the majority of daily physical phenomena. For deformable and granular objects, we utilize MPM implemented by Taichi \cite{TaichiLang}. Liquid simulations employ SPH through Doriflow \cite{Doriflow}. More details of simulation, including the ranges of various physical parameters, are provided in Appendix \ref{app:simulating}.

We note that while current simulators may not achieve perfect physical fidelity, their errors are thoroughly studied and controlled during simulation \cite{Macklin2016,Fatehi2011}. Thus, creating a massive synthetic dataset encompassing diverse physical phenomena unarguably holds significant potential to advance research progress, as also demonstrated by recent work \cite{Zhao2025a}.  

\subsection{Sampling Cameras and Rendering Videos}\label{sec:datset_videos}

For each 3D scene, we place 12 static cameras uniformly across the upper hemisphere at elevation angles $30^{\circ}\sim60^{\circ}$ to capture comprehensive multiview coverage, while a single moving camera records a challenging monocular video throughout the entire physical activity. Note that, the static cameras' positions vary across scenes according to different spatial arrangements and backgrounds. The moving camera follows predefined trajectories, typically circling the activity at random elevation angles. All videos are rendered at 1120$\times$1120 resolution with 30 FPS (60 FPS for 8780 scenes containing laser activities), averaging 5.2 seconds in duration to fully capture each physical activity. This results in 2 million dynamic videos. More details are in Appendix \ref{app:cameras_render}.

\subsection{Annotating and Splitting Dataset}\label{sec:dataset_labels}

For each 3D scene, we provide comprehensive annotations including a detailed paragraph describing visual elements and the physical activity. These texts are manually added and proofread using Qwen3 to eliminate grammatical errors, averaging about 64 English words per scene. During video rendering, we simultaneously generate: ground truth depth images, per-frame object masks, 3D trajectories of dynamic objects, object meshes, and material properties. Our annotations thus encompass five key aspects: geometry, semantics, motion, physical properties, and textual descriptions.

The dataset is partitioned into train/validation/held-out test sets using an 8:1:1 ratio. We ensure all 3D assets appear exclusively in one partition to prevent potential data leakage. Our dataset quality is ensured through a structured workflow including standardized development and independent validation. More details are in Appendix \ref{app:annotate_split}.

%% file: sec/4_applications.tex
\section{Applications}

\subsection{Physics-aware Video Generation}\label{sec:exp_vid_gen}

While video generation models demonstrate significant advancements in visual fidelity, they frequently fail to capture physically plausible dynamics \cite{Motamed2025,Kang2024}. \nickname{} contains diverse and massive dynamic activities following a variety of daily physical phenomena. This rich repository serves as an ideal resource for training or fine-tuning next-generation video models that faithfully emulate real-world physics.

In this section, we conduct fine-tuning experiments on three representative video models: 1) \textbf{SVD}-XT \cite{Blattmann2023}, a UNet-based image-to-video (I2V) model; 2) \textbf{CogVideoX}-1.5-5B \cite{Yang2025}, a Transformer-based text-image-to-video (TI2V) model; and 3) \textbf{Wan2.2}-5B \cite{Wan2025}, the latest Transformer and flow matching based TI2V model in the field. For these models, we adopt three commonly used fine-tuning techniques: 1) Low-Rank Adaptation (\textbf{LoRA}) \cite{Hu2021}, 2) Supervised Fine-Tuning (\textbf{SFT}) \cite{Howard2018}, and 3) Final Layer Tuning (\textbf{FLT}) \cite{Yosinski2014}. To showcase the potential of \nickname{}, we randomly sample a subset of training videos (83,650 text-video pairs) for fine-tuning all models until convergence. All fine-tuned models and their original models are then evaluated on a subset of test videos (772 text-video pairs, called \textit{test-small}). More details of experiment settings are provided in Appendix \ref{app:vid_gen_exp}.   

\textbf{Metrics}: Traditional metrics such as \textbf{FVD} \cite{Unterthiner2018} primarily assess visual realism in video generation, but are inadequate for evaluating the physical plausibility of motion. Some recent studies \cite{He2024,Bansal2025} use video-based VLMs to assess physical commonsense. However, these models often fail to produce meaningful evaluations, as they fundamentally lack an understanding of physical laws and are thus ill-suited for judging physical correctness. Other works \cite{Meng2025,Chow2025,Shen2025} introduce benchmarks featuring QA tasks to probe physical understanding, but they are typically qualitative and cannot quantitatively measure the correctness of physical motions. 

To this end, we introduce a novel metric for the quantitative assessment of physical motion fidelity in generated videos. Particularly, given a reference video $\mathcal{V}_{ref}$ exhibiting physically accurate motion (\eg{}, from our test set) and an AI-generated video $\mathcal{V}_{gen}$ 
produced using identical initial frame(s) and textual prompts as $\mathcal{V}_{ref}$, ensuring controlled comparison conditions, we apply discrete Fourier transform (DFT) to obtain their respective frequency domain representations. We then compare the energy (squared amplitude) of $DFT(\mathcal{V}_{ref})$ and $DFT(\mathcal{V}_{gen})$ and introduce \textbf{Physical Motion Fidelity (PMF)}, defined as follows:
\vspace{-0.1cm}
\begin{equation}
\setlength{\abovedisplayskip}{3pt}
\setlength{\belowdisplayskip}{3pt}
    PMF = f_{energy}\Big(
    DFT(\mathcal{V}_{gen}), DFT(\mathcal{V}_{ref})
    \Big)
\end{equation}

PMF quantifies kinematic discrepancies between the dynamic trajectories in $\mathcal{V}_{gen}$ and $\mathcal{V}_{ref}$. Crucially, higher PMF scores indicate less deviation from the reference motion patterns. The metric fundamentally differs from pixel-level similarity measures by evaluating physical fidelity rather than visual correspondence, as frame-perfect alignment is neither achievable nor desirable in generative tasks.

To empirically validate physical plausibility from a human perspective, we additionally conduct a user study to assess generated videos. Higher human ratings reflect greater perceived physical plausibility in the video content. More details of our metric and human rating are in Appendix \ref{app:vid_gen_exp}.

\begin{figure*}[t]
\centering
\includegraphics[width=1\linewidth]{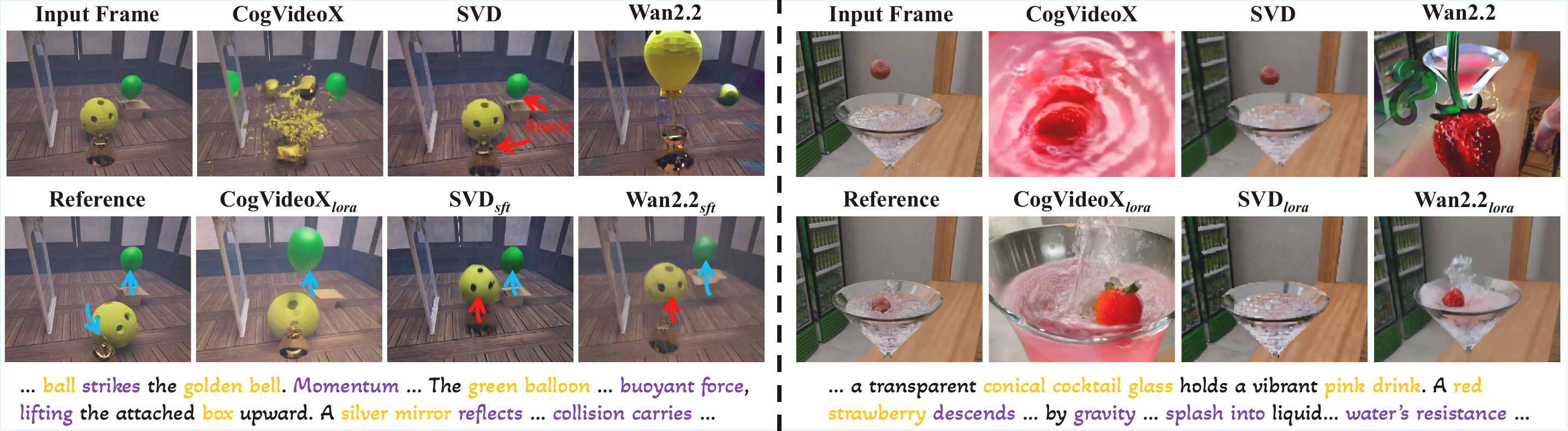}
\vskip -0.1in
\caption{Qualitative examples demonstrating improved physical plausibility in videos generated after fine-tuning on \nickname{}.}
\label{fig:vid_gen_res}
\vspace{-0.2cm}
%\vskip 0.4in
\end{figure*}

\begin{figure*}[t]
\centering
\includegraphics[width=1\linewidth]{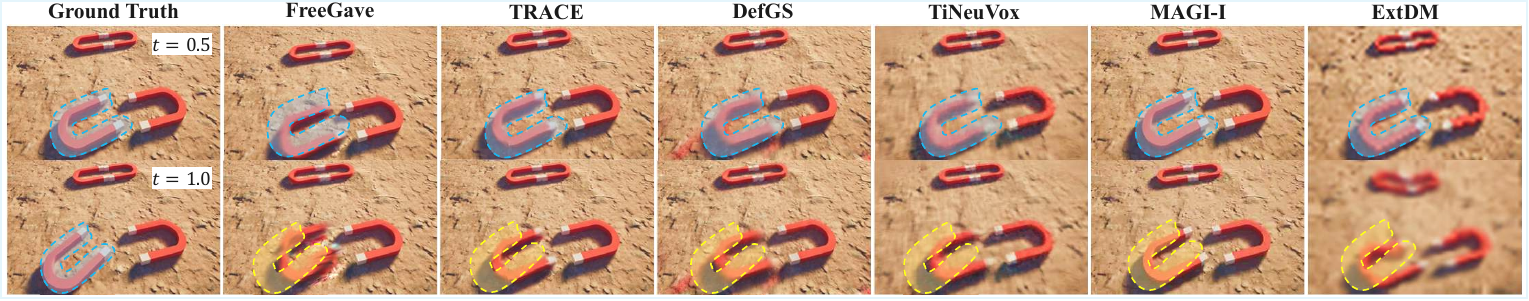}
\vskip -0.1in
\caption{Qualitative examples of long-term future frame prediction by current methods for trained viewpoints.}
\label{fig:future_pred_res}
\vspace{-0.4cm}
%\vskip 0.4in
\end{figure*}

\begin{table}[thb] \tabcolsep=0.3cm \vspace{-0.3cm}
\centering
\caption{The left part shows quantitative results of video generation models with and without fine-tuning on \nickname{}. The right part shows PMF scores across four physical domains separately.
}\vspace{-0.25cm}
\label{tab:vid_gen_res}
\setlength{\tabcolsep}{2.0pt}
\resizebox{0.48\textwidth}{!}{
\begin{tabular}{lccccccc}
\toprule[1.0pt]
\rowcolor{headergray} 
 & & & Human& Mechanics & Magnetism& Optics & Fluid \\

\rowcolor{headergray} 
& \multirow{-2}{*}{PMF $\uparrow$} & \multirow{-2}{*}{FVD $\downarrow$} & Rating $\uparrow$ & PMF $\uparrow$ & PMF $\uparrow$ & PMF $\uparrow$ &PMF $\uparrow$ \\ 

\toprule[1.0pt]
SVD \cite{Blattmann2023} & 2.753 & 203 & \textbf{6.09} & \cellcolor[HTML]{FFD8BF} 2.763  & \cellcolor[HTML]{FF7B27} 3.758 & \cellcolor[HTML]{FFE3D1} 2.582 & \cellcolor[HTML]{FFAF7C} 3.278\\
SVD$_{lora}$  & 2.446 & 150 & 5.82  & \cellcolor[HTML]{FFE9DB} 2.473  & \cellcolor[HTML]{FFAF7B} 3.283 & \cellcolor[HTML]{FFF0E7} 2.303 & \cellcolor[HTML]{FFD5BA} 2.815\\
SVD$_{sft}$   & \textbf{3.147} &  \textbf{143} & 6.08 & \cellcolor[HTML]{FFBC90} 3.139 & \cellcolor[HTML]{FF6400} 3.948 &  \cellcolor[HTML]{FFB17E} 3.261 & \cellcolor[HTML]{FF6E10} 3.868\\
SVD$_{flt}$   & 2.464 & 147 & 5.45  & \cellcolor[HTML]{FFE9DB} 2.463 & \cellcolor[HTML]{FFA064} 3.426 & \cellcolor[HTML]{FFE8DA} 2.485 & \cellcolor[HTML]{FFC29B} 3.061 \\ \hline
CogVideoX \cite{Yang2025}  & \textbf{2.877} & 165 & \textbf{2.98} &  \cellcolor[HTML]{FFCEAE} 2.912 & \cellcolor[HTML]{FFC5A0} 3.025 & \cellcolor[HTML]{FFE8D9} 2.495 & \cellcolor[HTML]{FFC49F} 3.033\\
CogVideoX$_{lora}$  &  2.869 & \textbf{149} & 2.95 & \cellcolor[HTML]{FFD0B2} 2.881 & \cellcolor[HTML]{FFC6A2} 3.008 & \cellcolor[HTML]{FFE8DA} 2.482 & \cellcolor[HTML]{FFB687} 3.206\\ \hline
Wan2.2-5B \cite{Wan2025}  & 2.041 & 258 & 2.26 & \cellcolor[HTML]{FFF9F6} 2.031 & \cellcolor[HTML]{FFD9C0} 2.752 & \cellcolor[HTML]{FFFFFF} 1.588 & \cellcolor[HTML]{FFB687} 3.205\\
Wan2.2-5B$_{lora}$   & 2.785 & \textbf{178} & 4.80 & \cellcolor[HTML]{FFD8C0} 2.760 & \cellcolor[HTML]{FFA46A} 3.390  & \cellcolor[HTML]{FFE4D3} 2.566 & \cellcolor[HTML]{FF802F} 3.716\\
Wan2.2-5B$_{sft}$  & \textbf{2.978} & 190 & \textbf{5.95} & \cellcolor[HTML]{FFC8A5} 2.990 & \cellcolor[HTML]{FF9D5E} 3.462 & \cellcolor[HTML]{FFCFB1} 2.888 & \cellcolor[HTML]{FFA971} 3.344 \\
Wan2.2-5B$_{flt}$  & 2.227 & 341 & 2.61 & \cellcolor[HTML]{FFF3EC} 2.227 & \cellcolor[HTML]{FFB585} 3.214 & \cellcolor[HTML]{FFFCFA} 1.908 & \cellcolor[HTML]{FFAB74} 3.325\\ \hline
\toprule[1.0pt]
\end{tabular}
}\vspace{-0.3cm}
\end{table}

The left part of Table \ref{tab:vid_gen_res} shows that SVD and Wan2.2 achieve notable improvements in PMF scores after fine-tuning, especially using SFT technique. This validates the effectiveness of \nickname{} in imbuing models with physical knowledge. We also observe that human perception of physical dynamics highly correlates with our newly introduced PMF metric. The right part of Table \ref{tab:vid_gen_res} shows that most models exhibit higher accuracy in magnetism and fluid, but lower scores in mechanics and optics, highlighting the challenges for future research in learning different types of physics. We will continue to benchmark the latest video generation models and update results on our \href{https://vlar-group.github.io/PhysInOne.html}{website}.

\subsection{Future Frame Prediction}
Accurate prediction of future frames necessitates a model's comprehension of underlying physical dynamics, with critical applications in autonomous driving, embodied AI, \etc{}. Input-output specifications may vary significantly in different applications. For example, in robot manipulation, given multiview videos as input, the model aims to predict precise dexterous control, where accurate and continuous short-term future predictions are essential for action execution. Conversely, video understanding tasks typically prioritize long-term predictions from monocular video inputs. Our collection of both multiview and monocular video data in \nickname{} enables demonstration of these diverse applications. This section introduces two tasks: long-term and short-term future frame prediction under varied settings.

\begin{figure*}[t]
\centering
\includegraphics[width=0.99\linewidth]{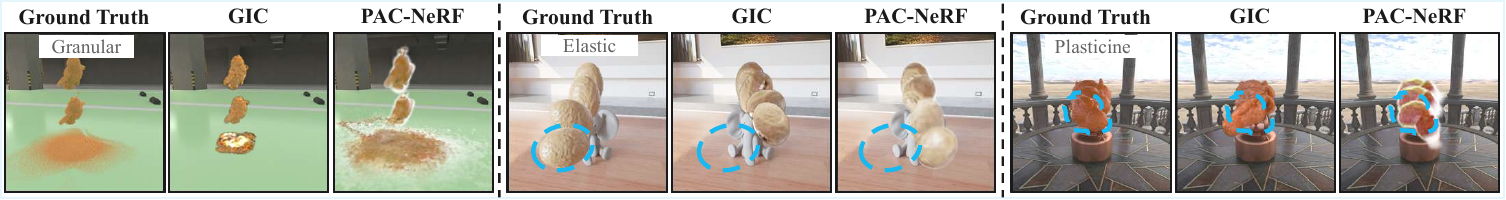}
\vskip -0.1in
\caption{Qualitative resimulation results using estimated physical properties. Both baselines fail to accurately infer properties for complex objects against intricate backgrounds, leading to physically implausible outcomes. }
\label{fig:phys_prop_res}
\vspace{-0.2cm}
%\vskip 0.4in
\end{figure*}

\subsubsection{Long-term Future Frame Prediction}
In this experiment, models are tasked with predicting the second half of a dynamic 3D scene ($\sim$2.6 seconds ahead, $\sim$78 future frames) from the test set, given the first half video clip as input. We evaluate the following two groups of existing methods for this task.
\begin{itemize}[leftmargin=*] %\vspace{-0.2cm}
    \setlength{\itemsep}{1pt}
    \setlength{\parsep}{1pt}
    \setlength{\parskip}{1pt}
        \item \textit{4D Modeling Methods}: \textbf{TiNeuVox} \cite{Fang2022}, \textbf{DefGS} \cite{Yang2024c}, \textbf{FreeGave} \cite{Li2025c}, and \textbf{TRACE} \cite{Li2025b}. These methods model dynamic 3D scenes from multiview videos using scene-specific deformation or velocity fields, enabling future frame prediction from arbitrary (seen and novel) views. 
        \item \textit{Video Prediction Methods}: \textbf{ExtDM} \cite{Zhang2024}, and \textbf{MAGI-1} \cite{AI2025}. 
        ExtDM is trained from scratch on randomly sampled 83,650 videos (same as used in Section \ref{sec:exp_vid_gen}), while MAGI-1 is evaluated using its pretrained model.
\end{itemize}

Given that training scene-specific models for the entire test set is neither feasible nor necessary, we evaluate all six methods on a randomly selected subset of 103 test scenes, called \textit{test-mini}. This assesses their long-term future prediction capabilities for both seen and novel viewing angles. We report standard per-frame metrics \textbf{PSNR}/\textbf{SSIM}/\textbf{LPIPS}, along with our proposed dynamics metric, \textbf{PMF}. More details of experiment settings are provided in Appendix \ref{app:future_pred_exp}.
\begin{table}[thb] \tabcolsep=0.15cm \vspace{-0.3cm}
\centering
\caption{Quantitative results of long-term future frame prediction from seen / novel viewpoints on \nickname{}.}\vspace{-0.25cm}
\label{tab:long_future_pred_res}
\resizebox{0.48\textwidth}{!}{
\begin{tabular}{lccccc}
\toprule[1.0pt]
\rowcolor{headergray} & PMF $\uparrow$    & PSNR $\uparrow$ & SSIM $\uparrow$ & LPIPS$\downarrow$   \\ \toprule[1.0pt]
TiNeuVox \cite{Fang2022} & 3.710 / 2.885  & 21.49 / 15.20 & 0.633 / 0.452 & 0.517 / 0.665   \\
DefGS \cite{Yang2024c} & 3.980 / 3.347 & 22.85 / 17.95 & 0.833 / 0.598 & 0.192 / 0.348   \\
TRACE\cite{Li2025b}  & 3.869 / 3.242 & 22.42 / 17.44 & 0.756 / 0.599 & 0.295 / 0.422    \\
FreeGave \cite{Li2025c} & 3.897 / 3.265 & 22.57 / 17.75 & 0.818 / 0.619 & 0.219 / 0.355  \\ \hline

ExtDM \cite{Zhang2024} & 3.363 / -  & 19.55 / - & 0.657 / - & 0.771 / -    \\
MAGI-1 \cite{AI2025} & 4.086 / - & 23.14 / - & 0.788 / - & 0.364 / -     \\ \hline
\toprule[1.0pt]
\end{tabular}
}\vspace{-0.3cm}
\end{table}

From Table \ref{tab:long_future_pred_res}, we can see that current methods demonstrate reasonable future frame prediction for trained viewpoints, but fail to maintain quality under novel viewpoints. This underscores the significant challenge of modeling complex physical motions in 3D space, a gap we expect our dataset will inspire more advanced methods to address. 
Figure \ref{fig:future_pred_res} shows qualitative results.

\subsubsection{Continuous Short-term Future Frame Prediction}
In this task, we emulate potential robotic manipulation scenarios by requiring models to continuously predict the next 10 frames in a real time manner, given a stream of continuously observed frames as input. For efficiency, we evaluate four methods \textbf{DefGS} \cite{Yang2024c}, \textbf{FreeGave} \cite{Li2025c}, \textbf{ExtDM} \cite{Zhang2024}, and \textbf{MAGI-1} \cite{AI2025} on the same set of 103 dynamic 3D scenes on both seen and novel viewing angles. Table \ref{tab:short_future_pred_res} shows the quantitative results. More details of experiment settings are provided in Appendix \ref{app:future_pred_exp}.
\begin{table}[thb] \tabcolsep=0.15cm  \vspace{-0.3cm}
\centering
\caption{Quantitative results of short-term future frame prediction from seen / novel viewpoints on \nickname{}.}\vspace{-0.25cm}
\label{tab:short_future_pred_res}
\resizebox{0.48\textwidth}{!}{
\begin{tabular}{lccccc}
\toprule[1.0pt]
\rowcolor{headergray}    & PMF $\uparrow$  & PSNR $\uparrow$ & SSIM $\uparrow$ & LPIPS$\downarrow$  \\ \toprule[1.0pt]
DefGS \cite{Yang2024c} & 4.536 / 3.728  & 26.02 / 20.92 & 0.861 / 0.739 & 0.206 / 0.322   \\
FreeGave \cite{Li2025c} & 4.742 / 3.706 & 27.09 / 20.80 & 0.876 / 0.715 & 0.199 / 0.336   \\ \hline

ExtDM \cite{Zhang2024} & 3.774 / - & 22.14 / - & 0.717 / - & 0.715 / -    \\
MAGI-1 \cite{AI2025} & 4.696 / -  & 26.75 / - & 0.886 / - & 0.116 / -    \\ \hline
\toprule[1.0pt]
\end{tabular}
}\vspace{-0.3cm}
\end{table}

\subsection{Physical Properties Estimation}

Estimating the physical properties of objects from visual frames is particularly challenging due to complex dynamics arising from internal forces that vary with object deformation, non-uniform velocity distributions, and spatially heterogeneous external forces. These factors contribute to highly intricate visual appearances. Existing methods are often evaluated on relatively small datasets with limited object diversity, constraining progress in the field. Our collection, encompassing a wide range of objects, including deformable, granular, and liquid types together with complex backgrounds, provides an ideal platform for testing the capabilities of current and future models.

In this section, we evaluate two representative and scene-specific methods \textbf{PAC-NeRF} \cite{Li2023} and \textbf{GIC} \cite{Cai2024} on a randomly selected subset of 20 scenes in our test set, called \textit{test-tiny}. The scenes are distributed across five representative material categories (4 scenes per category):

\begin{figure*}[t]
\centering
\includegraphics[width=1\linewidth]{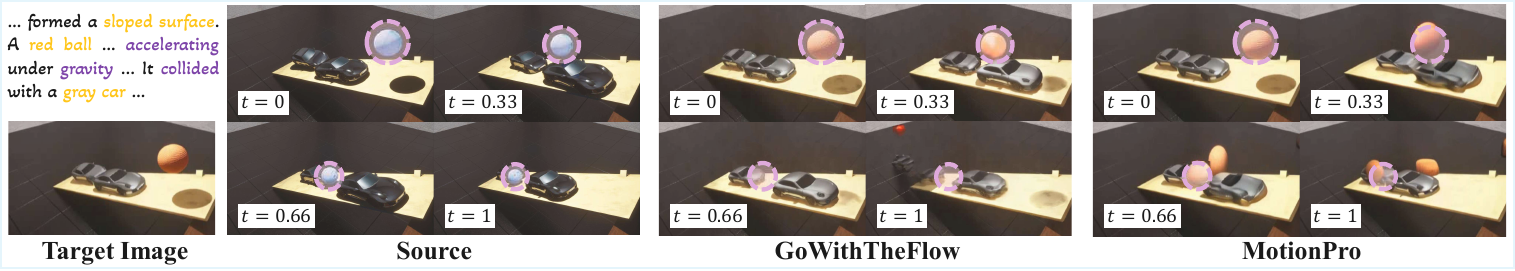}
\vskip -0.1in
\caption{Qualitative motion transfer results from GoWithTheFlow and MotionPro. Generated frames retain visual realism but fail to transfer complex physical motions (\eg{}, moving cars, falling ball).}
\label{fig:motion_transfer_res}
\vspace{-0.4cm}
%\vskip 0.4in
\end{figure*}

\begin{table}[tbp] \tabcolsep=0.12cm \vspace{-0.05cm}
\centering
\caption{Quantitative results of physical parameters estimation for five types of materials on \nickname{}. }\vspace{-0.3cm}
\label{tab:phys_prop_esti}
\resizebox{0.48\textwidth}{!}{%
\begin{tabular}{l|ccccc}
\toprule
\rowcolor{headergray}   & \multicolumn{5}{c}{Elastic Solids} \\
   & $\log_{10}(E)$ & $\nu$ & & & $\boldsymbol{v}$ \\ \hdashline
PAC-NeRF \cite{Li2023} & 117.18$\pm$68.44 & 14.26$\pm$7.94 & & & 4.04$\pm$1.11  \\
GIC \cite{Cai2024}    & 49.76$\pm$8.78 & 16.35$\pm$1.36 & & & 3.32$\pm$1.34  \\ \hline

\rowcolor{headergray}   & \multicolumn{5}{c}{Plasticine} \\
   & $\log_{10}(E)$ & $\nu$ & $\log_{10}(\tau_Y)$ & & $\boldsymbol{v}$ \\ \hdashline
PAC-NeRF \cite{Li2023} & 68.38$\pm$44.14 & 15.79$\pm$11.07 & 25.51$\pm$18.34 & & 3.25 $\pm$0.23  \\
GIC \cite{Cai2024}    & 178.36$\pm$46.68 & 42.72$\pm$25.65 & 17.11$\pm$19.38 & & 3.39$\pm$0.87  \\ \hline
\rowcolor{headergray}   & \multicolumn{5}{c}{Newtonian Fluids} \\
   & $\log_{10}(\mu)$ & $\log_{10}(\kappa)$ &  &  & $\boldsymbol{v}$ \\ \hdashline
PAC-NeRF \cite{Li2023} & 42.64$\pm$29.88 & 287.56$\pm$194.98 & & & 3.11$\pm$0.31  \\
GIC \cite{Cai2024}  & 8.78$\pm$9.64 & 70.07$\pm$53.44 &  & & 3.28$\pm$0.85  \\ \hline

\rowcolor{headergray}   & \multicolumn{5}{c}{Non-Newtonian Fluids} \\
   & $\log_{10}(\mu)$ & $\log_{10}(\kappa)$ & $\log_{10}(\tau_Y)$  & $\log_{10}(\eta)$ & $\boldsymbol{v}$ \\ \hdashline
PAC-NeRF \cite{Li2023} & 
309.42$\pm$235.61 & 552.89$\pm$105.24 &
339.20$\pm$145.99 &  65.60$\pm$71.73 & 2.95$\pm$0.30  \\
GIC \cite{Cai2024} & 
124.26$\pm$98.35 & 181.87$\pm$113.70 &
28.78$\pm$13.89 &  24.97$\pm$15.53 & 3.73$\pm$0.62  \\ \hline

\rowcolor{headergray}   & \multicolumn{5}{c}{Granular Substances} \\
   & $\theta_{fric}$ &  & &  & $\boldsymbol{v}$ \\ \hdashline
PAC-NeRF \cite{Li2023} & 16.87$\pm$27.36 & & & & 3.29$\pm$0.21  \\
GIC \cite{Cai2024}  & 18.85$\pm$16.67 & & & & 3.57$\pm$0.95  \\
\bottomrule
\end{tabular}
}\vspace{-0.3cm}
\end{table}

\begin{itemize}[leftmargin=*] %\vspace{-0.2cm}
\setlength{\itemsep}{1pt}
\setlength{\parsep}{1pt}
\setlength{\parskip}{1pt}
    \item \textit{Elastic Solids}: whose dynamics are governed by \textbf{Young’s modulus} $E$ and \textbf{Poisson’s ratio} $\nu$.
    \item \textit{Plasticine}: whose dynamics are governed by \textbf{Young’s modulus} $E$, \textbf{Poisson’s ratio} $\nu$, and \textbf{yield stress} $\tau_Y$.
    \item \textit{Newtonian Fluids}: whose dynamics are governed by \textbf{fluid viscosity} $\mu$ and \textbf{bulk modulus} $\kappa$.
    \item \textit{Non-Newtonian Fluids}: whose dynamics are governed by \textbf{yield stress} $\tau_Y$, \textbf{shear modulus} $\mu$, \textbf{plasticity viscosity} $\eta$, and \textbf{bulk modulus} $\kappa$.
    \item \textit{Granular Substances}: whose dynamics are governed by \textbf{friction angle} $\theta_{fric}$. 
\end{itemize}

\begin{table}[htb] \tabcolsep=0.3cm  \vspace{-0.3cm}
\centering
\caption{Quantitative results of resimulation based on estimated physical properties.}\vspace{-0.25cm}
\label{tab:phys_prop_esti_resim}
\resizebox{0.48\textwidth}{!}{
\begin{tabular}{lcccc}
\toprule[1.0pt]
\rowcolor{headergray} & PMF $\uparrow$    & PSNR $\uparrow$ & SSIM $\uparrow$ & LPIPS$\downarrow$    \\ \toprule[1.0pt]
PAC-NeRF \cite{Li2023} & 5.617 & 24.12  & 0.942 & 0.086    \\
GIC \cite{Cai2024} & \textbf{5.938} & \textbf{26.90}  & \textbf{0.950} & \textbf{0.074}     \\ \hline
\toprule[1.0pt]
\end{tabular}
}\vspace{-0.4cm}
\end{table}

We additionally estimate the initial velocity $\boldsymbol{v}$ of the dynamic object in each test scene. Table \ref{tab:phys_prop_esti} compares the accuracy of estimated physical properties against the ground truth provided in \nickname{}. To further validate the learned properties, we resimulate 3D scenes under novel initial conditions (\eg{}, modified object positions) and quantitatively compare rendered videos against reference videos generated using ground-truth physics under identical novel conditions in Table \ref{tab:phys_prop_esti_resim}. Figure \ref{fig:phys_prop_res} shows the qualitative results of resimulation. 
We can see that, while both models infer physically plausible properties, they fall short of accuracy in scenes with complex objects and backgrounds featured by our dataset. This demonstrates how our dataset's challenging cases provide critical benchmarks for revealing limitations in current physical reasoning capabilities. More details of experiment settings are provided in Appendix \ref{app:phys_esti_exp}.

\subsection{Motion Transfer}
Motion transfer seeks to propagate motion dynamics from a source video to a target image, synthesizing a new video that retains the target's visual attributes while adopting the source's motion patterns. Current approaches leverage optical flow to encode motion, achieving compelling results on datasets like DAVIS \cite{Pont-Tuset2017}. However, DAVIS primarily features simple motion patterns, such as a solitary swan gliding across water, with limited physical complexity. This raises a significant question about whether these methods can accurately transfer multiphysics interactions involving multiple objects from one video to another. This capability is essential for advanced applications like film and animation production, and virtual prototyping. 

This section evaluates recent methods \textbf{MotionPro} \cite{Zhang2025} and \textbf{GoWithTheFlow} \cite{Burgert2025} using a subset of 273 dynamic 3D scenes (source) from the \nickname{} val set. For quantitative assessment, we generate paired target scenes by replacing source objects with alternative shapes and materials while preserving identical physics activities. New videos rendered from these target scenes enable motion transfer performance evaluation. In Table \ref{tab:motion_transfer} and Figure \ref{fig:motion_transfer_res}, both methods maintain high visual fidelity overall, but fail to transfer intricate physical motions accurately, underscoring fundamental challenges in modeling complex physical dynamics, a gap we expect our dataset will motivate future methods to fill. More details of experiment settings are in Appendix \ref{app:motion_transfer_exp}.
\begin{table}[htb] \tabcolsep=0.3cm  \vspace{-0.3cm}
\centering
\caption{Quantitative results of transferring physical motions.}\vspace{-0.25cm}
\label{tab:motion_transfer}
\resizebox{0.48\textwidth}{!}{
\begin{tabular}{lccccc}
\toprule[1.0pt]
\rowcolor{headergray}  & PMF $\uparrow$   & PSNR $\uparrow$ & SSIM $\uparrow$ & LPIPS$\downarrow$   \\ \toprule[1.0pt]
GoWithTheFlow \cite{Burgert2025}  & 3.309 & 18.98  & 0.691 & \textbf{0.410}    \\
MotionPro \cite{Zhang2025} & \textbf{3.484} & \textbf{20.28}  & \textbf{0.775} & 0.467   \\ \hline
\toprule[1.0pt]
\end{tabular}
}\vspace{-0.4cm}
\end{table}

%% file: sec/5_conclusion.tex
\section{Conclusion} \label{sec:conclusion} We introduce \textbf{\nickname{}}, a comprehensive  dataset featuring 153,810 dynamic 3D scenes and 2 million videos covering 71 everyday physical phenomena. Experiments across four tasks demonstrate that fine-tuning on \nickname{} significantly improves the physical plausibility of foundation models. However, existing methods still struggle to accurately learn complex multiobject dynamics and physical properties, validating the dataset's value as a rigorous testbed.

%% file: sec/X_suppl.tex
\clearpage

\setcounter{page}{1}
% \maketitlesupplementary

\onecolumn
\begin{center}
        \Large
        \textbf{\thetitle}\\
        \vspace{0.5em}Supplementary Material \\
        \vspace{1.0em}
\end{center}

% \switchonecolumn

\subsection{More Details of Physical Phenomena and Laws}\label{app:phys_phen_laws}

We identify 71 basic physical phenomena commonly observed in everyday life. Each phenomenon involves one or more underlying physical laws, encompassing areas such as mechanics, optics, fluid dynamics, and magnetism. Table \ref{tab:list-physics-phen-law} lists these 71 phenomena along with the corresponding physical principles involved.

\begin{longtable}{r|m{8cm}|m{7cm}}
\caption{The list of Physical Phenomena and Related Physical Laws} \vspace{-0.2cm}\label{tab:list-physics-phen-law} \\
\hline
\rowcolor{headergray} \textbf{Index} & \textbf{Physical Phenomena} & \textbf{Physical Laws} \\ \hline
\endfirsthead

\hline
\rowcolor{headergray} \textbf{Index} & \textbf{Physical Phenomena} & \textbf{Physical Laws} \\ \hline
\endhead

\hline
\endfoot

\hline
\endlastfoot

1 & Object collide with static, stationary objects & Laws of Momentum \\ \hline
2 & Moving objects collide with non-static stationary objects & Laws of Momentum \\ \hline
3 & Two moving objects collide & Laws of Momentum \\ \hline
4 & Objects in equilibrium of wind and gravity & Equilibrium, Aerodynamics, Gravity\\ \hline
5 & Wind applied to a stationary object & Aerodynamics \\ \hline
6 & Wind applied to objects moving in same direction & Aerodynamics \\ \hline
7 & Wind applied to objects moving in the opposite direction & Aerodynamics \\ \hline
8 & Wind applied to moving objects changes its velocity (applied at an angle) & Aerodynamics \\ \hline
9 & Object thrown up with angle & Gravity \\ \hline
10 & Objects falling straight down & Gravity \\ \hline
11 & Objects rolling down a straight slope & Gravity, Friction \\ \hline
12 & Objects rolling up a slope & Gravity, Friction \\ \hline
13 & Magnetic Attraction & Magnetism \\ \hline
14 & Magnetic Repulsion & Magnetism \\ \hline
15 & Objects near uniformly rotating pillar & Laws of Rotation \\ \hline
16 & Objects near acceleratingly rotating pillar & Laws of Rotation \\ \hline
17 & Objects inside uniformly rotating bowl & Equilibrium, Laws of Rotation, Gravity, Friction \\ \hline
18 & Objects inside acceleratingly rotating bowl & Equilibrium, Laws of Rotation, Gravity, Friction \\ \hline
19 & Objects on uniformly rotating plane & Laws of Rotation, Friction \\ \hline
20 & Objects on acceleratingly rotating plane & Laws of Rotation, Friction\\ \hline
21 & Objects on coarse surface with friction & Friction \\ \hline
22 & Spring is compressed & Laws of Elasticity \\ \hline
23 & Spring is stretched & Laws of Elasticity \\ \hline
24 & Breakable object shatters & Laws of Plasticity \\ \hline
25 & Mirror shatters & Laws of Plasticity, Law of Reflection \\ \hline
26 & Elastic rope connection & Laws of Elasticity \\ \hline
27 & Object bounces off spring board & Laws of Elasticity, Laws of Momentum \\ \hline
28 & Objects interact with a balanced seesaw & Laws of Torque \\ \hline
29 & Objects interact with an imbalanced seesaw & Laws of Torque \\ \hline
30 & Free balloon floats to ceiling & Laws of Buoyancy \\ \hline
31 & Tethered balloon pulls string taut & Laws of Buoyancy, Rope Restraint
 \\ \hline
32 & Multiple balloons lifting & Laws of Buoyancy, Rope Restraint, Gravity \\ \hline
33 & Laser hits flat mirror and reflects & Law of Reflection \\ \hline
34 & Laser reflects off multiple mirrors & Law of Reflection \\ \hline
35 & Laser hits and reflects off concave mirror & Law of Reflection \\ \hline
36 & Laser hits and reflects off convex mirror & Law of Reflection \\ \hline
37 & Mirror sweeps beam & Law of Reflection \\ \hline
38 & Laser blocked by object & Light Obstruction \\ \hline
39 & Mirror reflection & Law of Reflection \\ \hline
40 & Cart moving forward with rolling wheels & Complex Mechanical Structure Constraints \\ \hline
41 & Objects on rotating turntable flies off & Laws of Rotation, Friction, Laws of Inertia \\ \hline
42 & Rotating block pushes another objects & Laws of Inertia, Laws of Rotation, Laws of Momentum \\ \hline
43 & One object carrying another & Laws of Inertia, Friction \\ \hline
44 & Catapult launches objects & Special Mechanical Structure, Gravity \\ \hline
45 & Chain suspends objects &  Complex Mechanical Structure Constraints, Gravity \\ \hline
46 & Objects Swing & Laws of Pendulum Motion, Gravity \\ \hline
47 & Double Pendulum Moves & Laws of Multiple Pendulum Motion, Gravity \\ \hline
48 & Crank push objects & Special Mechanical Structure \\ \hline
49 & Wall composed of square blocks collapses & Gravity, Structural Stability \\ \hline
50 & Wooden board supported by sticks collapses & Gravity, Structural Stability \\ \hline
51 & Objects float on the fluid surface & Laws of Buoyancy, Fluid Dynamics \\ \hline
52 & Objects drop into the fluid & Laws of Buoyancy, Fluid Dynamics \\ \hline
53 & Objects' movement causes fluid motion & Fluid Dynamics \\ \hline
54 & Flowing fluid carries objects along & Laws of Buoyancy, Fluid Dynamics \\ \hline
55 & Fluid flows against stationary objects & Fluid Dynamics \\ \hline
56 & Fluid transfers from one container to another & Fluid Dynamics, Conservation of Mass, Surface Tension \\ \hline
57 & Fluid passes through several connected containers & Fluid Dynamics, Conservation of Mass, Surface Tension \\ \hline
58 & Fluid flows through grid-like structures & Fluid Dynamics, Conservation of Mass \\ \hline
59 & Fluid moves across mountainous or uneven landscapes & Fluid Dynamics \\ \hline
60 & Increasing fluid volume elevates the surface level & Fluid Dynamics \\ \hline
61 & Fluid flows along the contours of an object's surface & Fluid Dynamics \\ \hline
62 & Jet-like fluid projection upward or outward & Fluid Dynamics \\ \hline
63 & Fluid exhibits surface tension &  Fluid Dynamics, Laws of Surface Tension\\ \hline
64 & Fluid refracts light when crossing media &  Fluid Dynamics, Optics, Law of Refraction (Snell’s Law)\\ \hline
65 & Sticky fluid drips and accumulates on objects &  Laws of Cohesion, Viscous Flow\\ \hline
66 & Sticky fluid falls from an object's surface & Laws of Cohesion, Laws of Viscous Flow (Navier-Stokes) \\ \hline
67 & An elastic object falls and bounces on another surface & Laws of Elasticity\\ \hline
68 & A plasticine object falls and deforms on a surface & Laws of Plasticity \\  \hline
69 & A Newtonian fluid falls and spreads across a surface & Laws of Viscous Flow (Navier-Stokes) \\ \hline
70 & A Non-Newtonian fluid falls and flows with variable resistance & Laws of Viscoplastics Flow\\ \hline
71 & A granular substance falls and disperses across a surface & Laws of Friction\\ \hline

\end{longtable}
% \end{strip}

\clearpage

\switchtwocolumn

\subsection{More Details of Collecting 3D Assets}\label{app:3d_assets}
To construct our dataset, we source 3D assets, including meshes and textures, from publicly available platforms such as SketchFab \cite{Sketchfab}, Fab \cite{FAB}, ShareTextures 
%(https://www.sharetextures.com), 
and Blender Kit. All downloaded assets have been verified to carry licenses compatible with non-commercial use, including CC BY-NC, CC BY-SA, CC BY-NC-SA, CC0, CC BY and RF. We ensure that every file complies with the licensing terms and can be legally used for building a non-commercial dataset.

For assets obtained from SketchFab, we apply a filtering process to select meshes that meet the required licensing conditions and have verified that the authors did not explicitly prohibit AI-related usage. 
\begin{figure*}[ht!]
\centering
\includegraphics[width=0.9\linewidth]{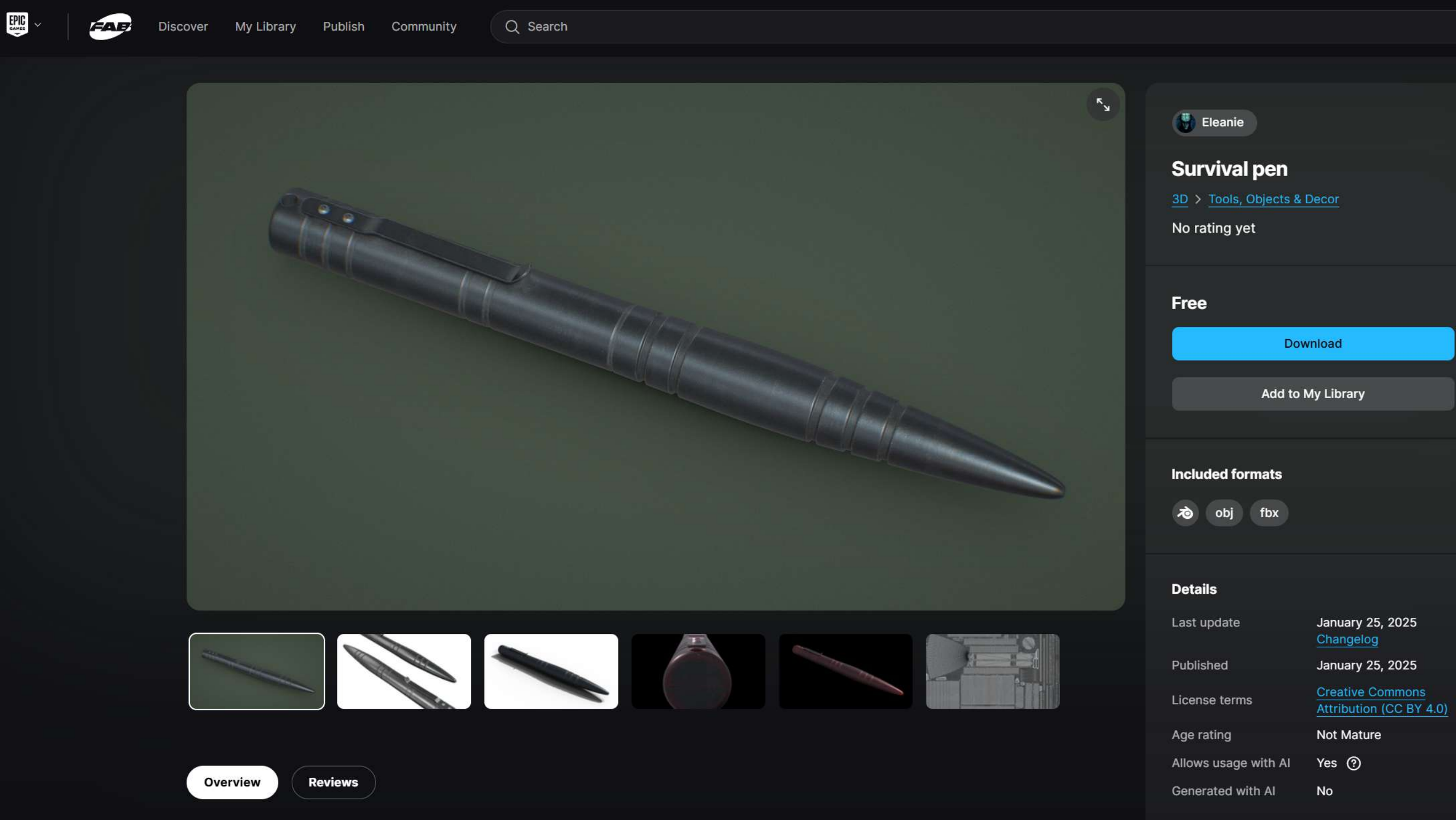}
\vskip -0.1in
\caption{Exemplary 3D asset under CC BY License}
\label{fig:Fab_CCBY}
\vspace{-0.2cm}
%\vskip 0.4in
\end{figure*}

\begin{figure*}[ht!]
\centering
\includegraphics[width=0.9\linewidth]{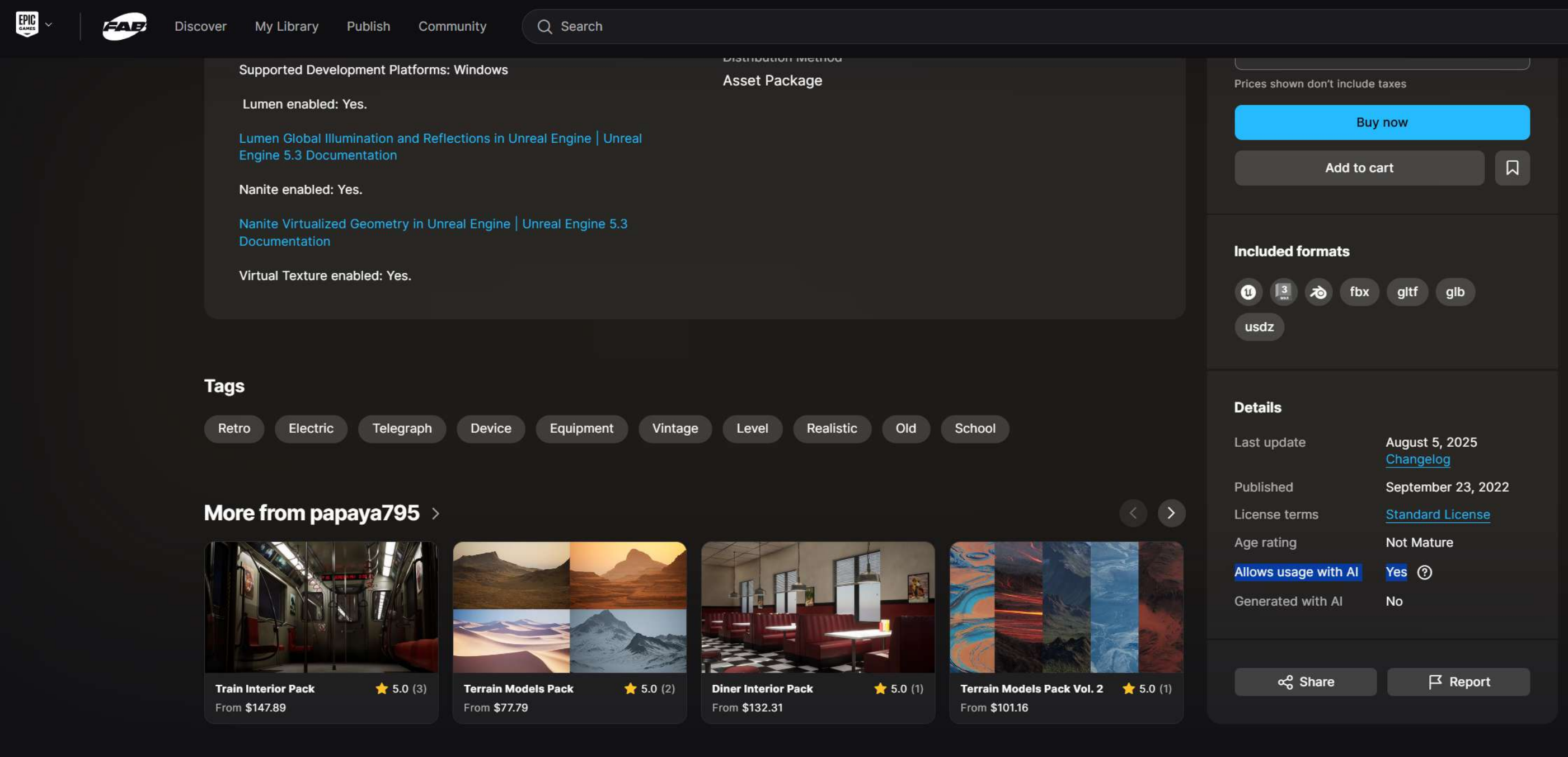}
\vskip -0.1in
\caption{Exemplary 3D asset under UE Standard License}
\label{fig:Fab_standard}
\vspace{-0.2cm}
%\vskip 0.4in
\end{figure*}
For assets downloaded from Fab, we ensure that they are licensed under either the CC BY License or the Unreal Engine Standard License, and that the authors explicitly stated these assets can be used for AI-related purposes. Figure~\ref{fig:Fab_CCBY} and Figure~\ref{fig:Fab_standard} respectively show examples of assets under the CC BY License and the Unreal Engine Standard License, with clear indications of \textit{Allows usage with AI}.

Assets from Blender Kit are distributed under a Royalty-Free (RF) license, which permits commercial use without requiring attribution. 
Our work complies with these terms as we are building a non-commercial, open-source dataset. 

\begin{figure*}[t!]
\centering
\includegraphics[width=1.\linewidth]{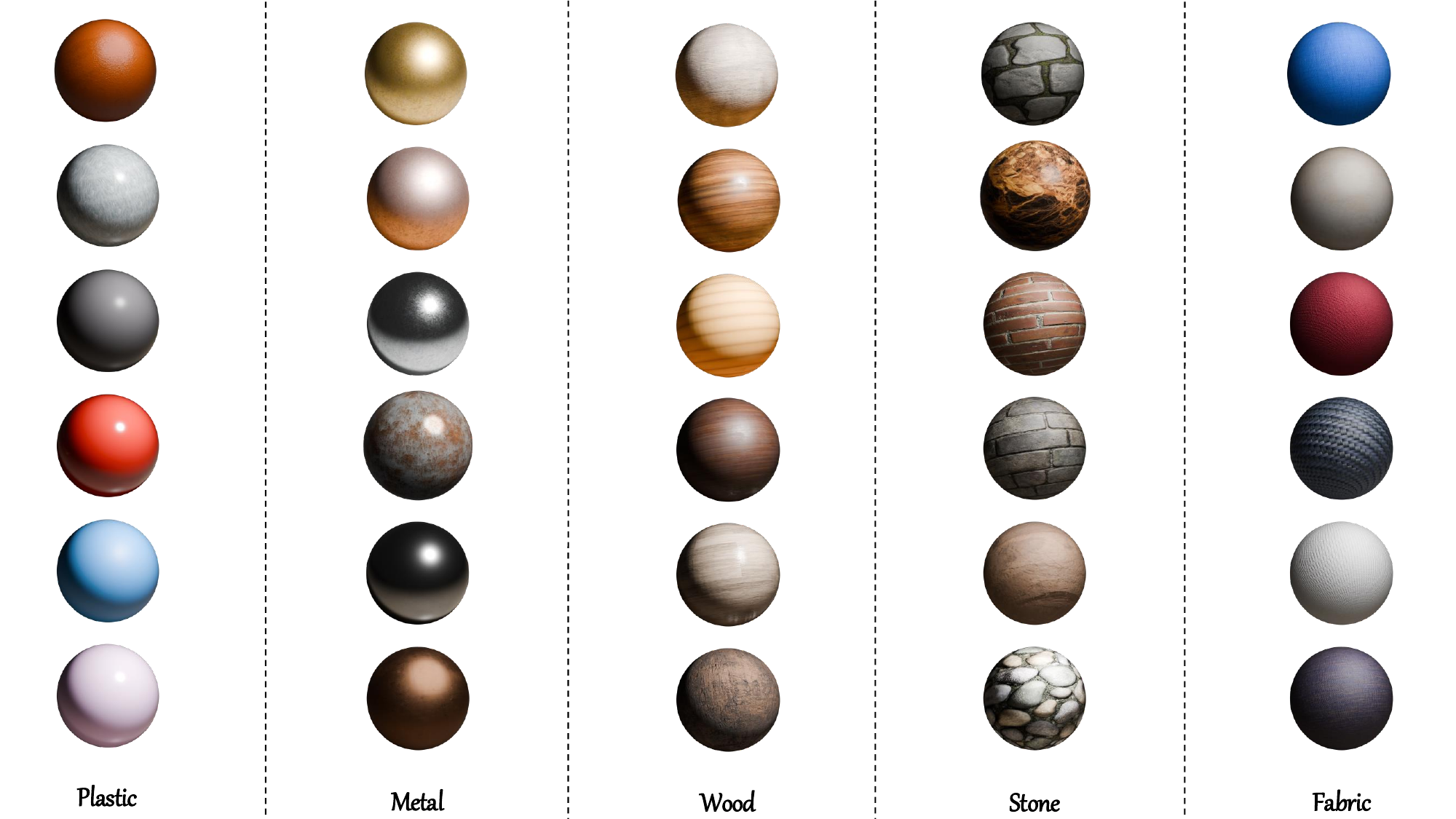}
\vskip -0.1in
\caption{Examples of materials.}
\label{fig:all_material}
\vspace{-0.2cm}
%\vskip 0.4in
\end{figure*}

Additionally, ShareTextures offers a wide range of free textures under the CC0 license. We incorporate textures selected from this platform and convert them into materials compatible with Unreal Engine. Furthermore, we supplement these with materials sourced from Fab, resulting in five major material categories comprising a total of 623 materials: metal, stone, wood, fabric, and plastic. Figure~\ref{fig:all_material} presents representative materials from each category.

Interactable Objects: The downloaded meshes are typically solid and lack joint-like structures, making them non-interactive even if they appear to have multiple movable parts. Common examples include swivel chairs, doors, windows, seesaws, and pendulums-objects frequently encountered in daily life. To transform a downloaded mesh from a completely solid object into an interactive one with movable joints, we follow several steps, leveraging Unreal Engine’s Physical Constraint tool to define relationships between parts.
Taking a swivel chair as an example: 1) We first use Unreal Engine’s modeling tools to split the chair model into two parts: the base and the seat. 2) We then create a Chair Blueprint, in which we define two static mesh components, base and seat, and then add a Physical Constraint to establish the relationship between them. Unreal Engine’s Physical Constraint supports four types of constraints: linear limit, linear motor, angular limit, and angular motor. 
\begin{itemize}
    \item \textbf{Linear Limit}: It restricts translational movement along one or more axes within a specified range.
    \item \textbf{Linear Motor}: It applies a constant or target-driven force to drive linear motion along an axis.
    \item \textbf{Angular Limit}: It constrains rotational movement around one or more axes within defined angular bounds.
    \item \textbf{Angular Motor}: It generates torque to achieve or maintain a target angular velocity or position.
\end{itemize}

Destructible Objects: In Unreal Engine, these objects are assets that can break apart or deform dynamically during physical activities, simulating realistic physical destruction. This feature enhances immersion by allowing objects such as walls, barrels, or glass panes to respond to forces like explosions, collisions, or player interactions. Unreal Engine implements destructible objects through NVIDIA’s Apex Destruction module, which uses fracture meshes and physics simulation to create breakable assets. 
Creating a destructible object in Unreal Engine typically follows this pipeline:
\begin{enumerate}
    \item \textbf{Prepare the Base Mesh}: To import a closed, manifold static mesh into Unreal Engine.
    \item \textbf{Create a Destructible Mesh}: In the Content Browser, right-click the static mesh and select \textit{Create Destructible Mesh}.
    \item \textbf{Apply Fracture Settings}: In the Destructible Mesh Editor, set chunk count, random seed, and depth layers for hierarchical destruction.
    \item \textbf{Configure Damage Properties}: Define damage threshold, impact damage, and enable debris timeout for performance.
\end{enumerate}

\begin{figure*}[t!]
\centering
\includegraphics[width=1.\linewidth]{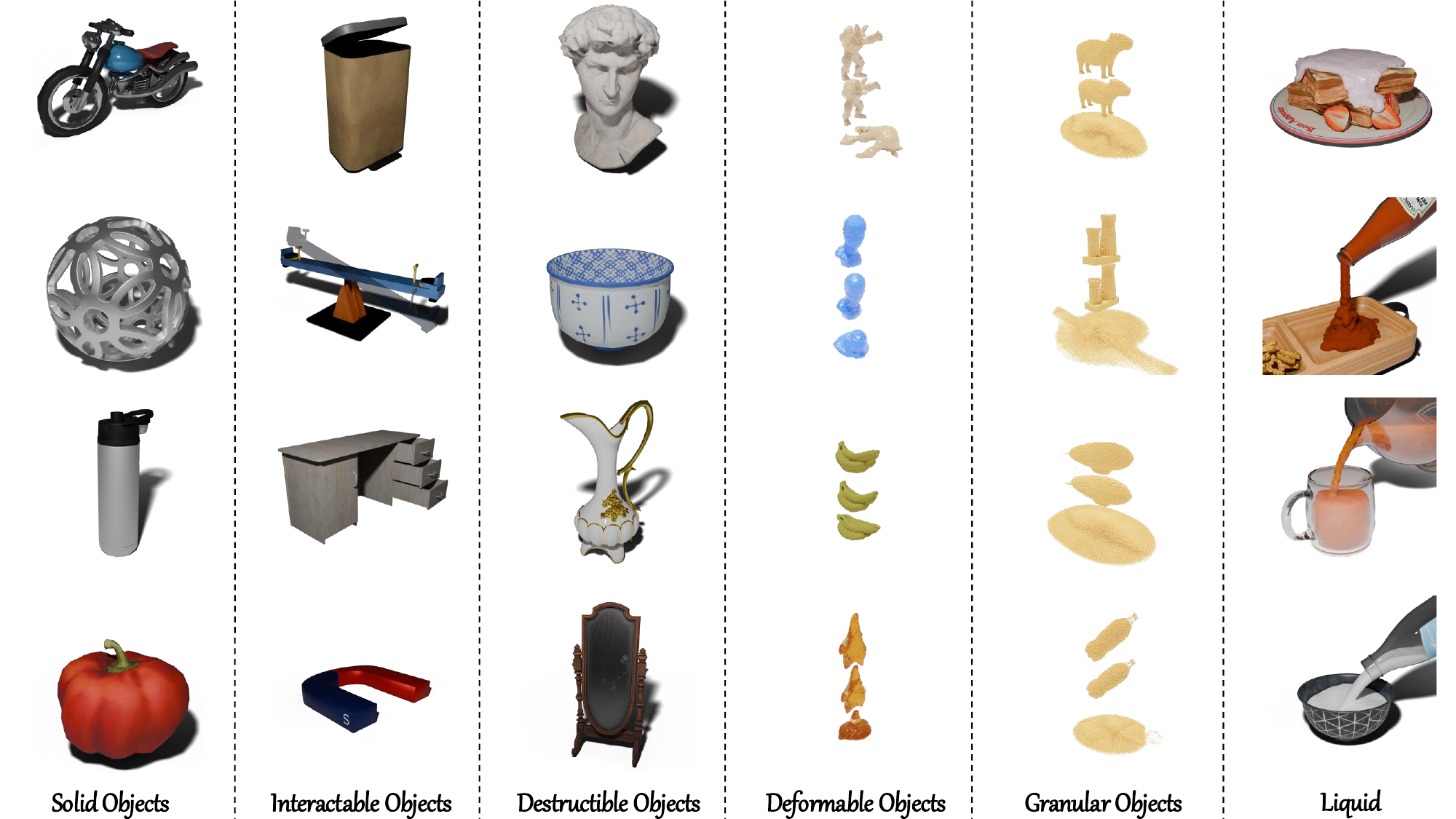}
\vskip -0.1in
\caption{Examples of 3D assets.}
\label{fig:all_mesh}
\vspace{-0.4cm}
%\vskip 0.4in
\end{figure*}

Figure~\ref{fig:all_mesh} showcases examples of objects. 

\subsection{More Details of Creating 3D Scenes}\label{app:3d_scenes}
In this section, we will describe how we create 3D dynamic scenes utilizing multiple physical engines.

\subsubsection{Creating 3D Scenes in Unreal Engine}

\begin{figure*}[t!]
\centering
\includegraphics[width=1.\linewidth]{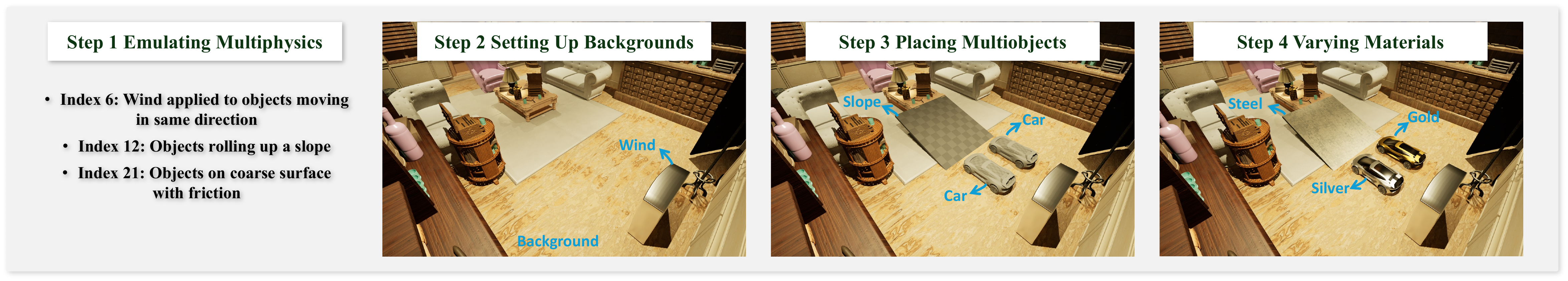}
\vskip -0.1in
\caption{The pipeline to create 3D scenes in Unreal Engine.}
\label{fig:UE_pipeline}
\vspace{-0.2cm}
%\vskip 0.4in
\end{figure*}

As an example, we illustrate the construction of a \textit{triple-physics activities} scenario. The pipeline is illustrated in Figure~\ref{fig:UE_pipeline}.

\noindent \textbf{Step 1: Emulating Multiphysics}

We randomly select three physical phenomena from Table~\ref{tab:list-physics-phen-law}. For example:
\begin{itemize}
    \item Index 6: Wind applied to an object moving in the same direction;
    \item Index 12: Object rolling up a slope;
    \item Index 21: Object on a coarse surface with friction.
\end{itemize}

These descriptions are intentionally abstract and serve as guidelines for Step~2 (background setting) and Step~3 (object placement), ensuring that the final setup incorporates all three phenomena.

\noindent \textbf{Step 2: Setting Up Backgrounds}

Based on the selected phenomena, we choose an indoor background to facilitate the construction of a slope and the placement of a fan, naturally aligning with an indoor setting.

\noindent \textbf{Step 3: Placing Multiobjects}

We position a wooden plank on the edge of a table to form a slope. Two small cars with initial velocities are placed at the bottom so they can roll upward. A fan is placed behind one car to apply wind force during motion.

\noindent \textbf{Step 4: Varying Materials}

To create more 3D scenes, we modify the slope to have a steel appearance and adjust its physical parameters. Similarly, the two cars are assigned different visual appearances and distinct physical properties.

\subsubsection{Creating 3D Scenes Concerning Liquid}
\begin{figure*}[t!]
\centering
\includegraphics[width=1.0\linewidth]{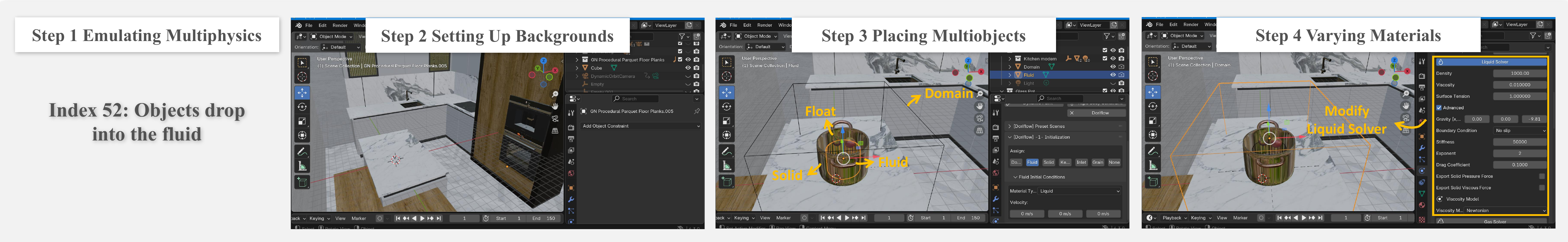}
\vskip -0.1in
\caption{The pipeline to create 3D scenes concerning liquid.}
\label{fig:doris_dataset_pipeline}
\vspace{-0.2cm}
%\vskip 0.4in
\end{figure*}

To construct a 3D scene with liquid, we follow the same four-step procedural pipeline. 
The pipeline is illustrated in Figure~\ref{fig:doris_dataset_pipeline}.

\noindent \textbf{Step 1: Emulating Multiphysics}
    
For liquid simulation, we include single-physics, double-physics, and triple-physics activities. We randomly select one to three physical phenomena from Table~\ref{tab:list-physics-phen-law}.
    
\noindent \textbf{Step 2: Setting Up Backgrounds}
    
One of 207 candidate backgrounds is selected. These provide contextual geometry, including diverse categories, such as bathroom, kitchen, \etc{}.

\noindent \textbf{Step 3: Placing Multiobjects} 
    
Multiple objects are placed against the background. We set objects that will not be driven by the liquid as solid (\eg{}, the container of the fluid), while objects that can be moved by the liquid are set as float. For a solid object, a domain will be set up to define the area for liquid simulation. 
To realize cases like stirring or pouring, we also design animations through key frames. 
Liquid can be initialized in arbitrary shapes, mainly depending on the shape of the liquid container. 
To enhance diversity, Inlets are also incorporated into the initialization (\eg{}, fountains).

\noindent \textbf{Step 4: Varying Materials} 
    
We select different fluid types, including Newtonian, non-Newtonian fluids, \etc{}. We adjust various physical parameters, such as viscosity, surface tension, \etc{}.

\subsubsection{Creating 3D Scenes Concerning Special Materials}
\begin{figure*}[t!]
\centering
\includegraphics[width=1.\linewidth]{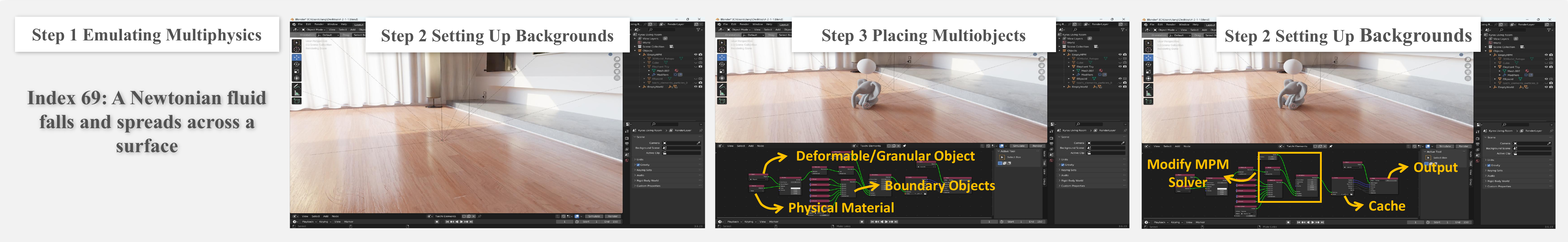}
\vskip -0.1in
\caption{The pipeline to create 3D scenes concerning special materials.}
\label{fig:mpm_dataset_pipeline_app}
\vspace{-0.2cm}
%\vskip 0.4in
\end{figure*}

Our dataset includes a subset of 1,440 dynamic 3D scenes designed to evaluate current methods for physical property estimation. 
This subset spans five representative material categories, elastic solids, plasticine, Newtonian fluids, non-Newtonian fluids, and granular substances, where each category is paired with distinct initial geometries to capture structural and behavioral variability. 
To construct each dynamic 3D scene, we follow the same four-step procedural pipeline.
The pipeline is illustrated in Figure~\ref{fig:mpm_dataset_pipeline_app}.

\noindent \textbf{Step 1: Emulating Multiphysics} 
    
We primarily capture the physical interactions of objects of different materials as they are released and collide under gravity, or as they fall onto boundary objects. We first randomly select one to three physical phenomena from the table \ref{tab:list-physics-phen-law}. For example:
\begin{itemize}
    \item Index 69: A Newtonian fluid falls and spreads across a surface.
\end{itemize}
    
\noindent \textbf{Step 2: Setting Up Backgrounds}
    
Then we will choose a background. These provide contextual geometry (\eg{}, walls, floors, or distant scenery) but they are purely passive, that they neither participate in nor perturb the underlying physics.

\noindent \textbf{Step 3: Placing Multiobjects}  
    
Objects are arbitrarily placed in the scene and defined as either boundary objects(static colliders) or the main object(deformable body) in the MPM simulation. These boundary objects have simple or complex geometry and directly contact with the main object, thus influencing local physical interactions based on boundary conditions. The main object is initialized as one of several candidate shapes, \eg{}, cube, sphere, torus, complex mesh, \etc{}.

\noindent \textbf{Step 4: Varying Materials}  

The materials of the main objects are selected from the five categories, which determine their physical behavior, including deformation, flow, and response to contact.

\subsection{More Details of Simulating Physical Dynamics}\label{app:simulating}

\subsubsection{Chaos Physics}
Unreal Engine’s Chaos Physics system is employed to simulate physically accurate dynamics within the created 3D scenes.  
It provides deterministic and high-performance simulation for complex interactions. Specifically, in our dataset, Chaos is responsible for:

\begin{itemize}
    \item \textbf{Rigid Body Dynamics}: Simulating the motion and interaction of solid objects under forces such as gravity, collisions, and impulses. This includes accurately handling of linear and angular momentum for static and dynamic objects.
    \item \textbf{Fracture and Destruction}: Managing breakable geometry through procedural fracturing and hierarchical destruction. Objects are divided into chunks that respond to impact forces, enabling realistic breakage and debris behavior during collisions or explosions.
\end{itemize}

Beyond these core functionalities, several interactive physical effects are implemented using Unreal Engine components combined with Chaos for accurate responses:

\begin{itemize}
    \item \textbf{Wind Force}: The wind effect is implemented as a rectangular force field region within the scene. Inside this region, a constant force direction is maintained, simulating a uniform wind flow. However, the force magnitude decays quadratically with the distance from the wind source, ensuring that objects closer to the origin experience stronger wind influence. At the far end of the force field, the force magnitude approaches zero, creating a realistic attenuation effect across the region. Chaos Physics computes the rigid body response to this spatially varying force distribution. 
    
    \item \textbf{Laser Interaction}: Laser beams are represented as meshes for visualization. Collision detection is handled by Chaos Physics to determine intersections with scene objects. Upon impact, the surface normal is retrieved, and the reflection direction is computed based on the law of reflection. Reflection occurs only on materials designated as mirrors, enabling controlled optical behavior. 
    
    \item \textbf{Magnetic Force}: We model magnetic interactions using a physics-based dipole field formulation. Specifically, the normalized magnetic vector field \(\mathbf{B}(\mathbf{p})\) generated by a source magnet's poles (\(\mathbf{p}_N, \mathbf{p}_S\)) at any spatial point \(\mathbf{p}\) is established as \(\mathbf{B}(\mathbf{p}) = \frac{\mathbf{r}_N}{\|\mathbf{r}_N\|^3} - \frac{\mathbf{r}_S}{\|\mathbf{r}_S\|^3}\), where \(\mathbf{r}_{N,S} = \mathbf{p} - \mathbf{p}_{N,S}\). 

When a target magnet enters this continuous vector field, the system samples \(\mathbf{B}\) at its respective poles to calculate the localized forces acting on each point. By aggregating these pole-specific forces, we analytically derive the translational force and rotational torque applied to the object. This formulation yields physically accurate kinematics, including homopolar repulsion, heteropolar attraction, and automatic streamline alignment. 
\end{itemize}

\subsubsection{Doriflow}

We employ Doriflow\cite{Doriflow}, a Blender add-on for fluid simulation, to produce physically consistent fluid sequences for our dataset. We use Blender 4.3.0 and Doriflow 1.3. We introduce a subset of 300 scenes for liquid simulation, mainly focusing on Newtonian and non-Newtonian fluid. 
 
Following the official Doriflow workflow, our simulations rely solely on explicitly defined geometry, emitters, and physical parameters. Assets of a 3D scene are imported into Blender and designated as either fluid emitters or solid colliders. Doriflow then performs the required preprocessing like generating internal voxel representations, preparing the simulation domain according to the user-specified resolution and bounds, and generating SPH particles, \etc{}. 

During simulation, Doriflow advances fluid states using its built-in GPU-accelerated solver. Each frame is computed through standard fluid-dynamics substeps, including advection, neighbor search, viscosity integration, external force application (\eg{}, gravity), pressure solving, and collision handling against the prepared collider geometry. The pressure and viscous forces involved in these steps are calculated as follows:

\begin{itemize}
    \item \textbf{Pressure Force Computation}:
      \[
        \mathbf{F}_{\text{press},i}=-\sum_{j}m_im_j\frac{p_i + p_j}{2\rho_i\rho_j}
\nabla W(\mathbf{r}_{ij},h)
      \]
      
      This equation calculates the pressure force on particle $i$ due to particle $j$ using the gradient of the smoothing kernel, where $\mathbf{F}_{\text{press},i}$ is the pressure force on particle $i$, $m_i$ and $m_j$ are masses, $p_i$ and $p_j$ are pressures, $\rho_i$ and $\rho_j$ are densities, $\nabla W$ is the gradient of the smoothing kernel, $\mathbf{r}_{ij}$ is the distance vector, and $h$ is the smoothing length.
      
    \item \textbf{Viscous Force Computation}:
      \[
        \mathbf{F}_{\text{visc},i}=\sum_{j}\frac{\mu}{2}
        \left(\frac{\mathbf{v}_i - \mathbf{v}_j}{\rho_i+\rho_j}\right)\cdot
\nabla^2 W(\mathbf{r}_{ij},h)
      \]
      
      This expression quantifies the viscous dissipation between particles, with $\mathbf{F}_{\text{visc},i}$ representing the viscous force acting on particle $i$, $\mu$ denoting the dynamic viscosity coefficient, $\mathbf{v}_i$ and $\mathbf{v}_j$ representing the velocity vectors, and $\nabla^2 W$ indicating the Laplacian of the smoothing kernel.
\end{itemize}

Fluid injection follows the emission rules defined at setup, enabling controlled generation of diverse flow behaviors.

After simulation completes, Doriflow converts the internal fluid representation into surface meshes for every frame through its built-in meshing stage. These meshes are stored directly as Blender geometry tracks and can be rendered or exported without further processing.

Doriflow allows configuration of parameters such as simulation resolution and time step size. For objects with key frames, Doriflow simulates the fluid based on the motion of these meshes. To enhance the reality, for large scenes such as mountains and valleys, we also add white water effects in Doriflow, including bubbles and foams. We adjust the transparency and size to improve the visual quality.

This integrated workflow provides a reproducible, Blender-centric method for generating high-fidelity fluid motion suitable for large-scale datasets. 

\subsubsection{Taichi Lang}\label{sec:taichi_lang}

We implement a full Material Point Method (MPM) simulation pipeline using Taichi and all constitutive and projection models in our solver, following the definitions provided in PAC-NeRF \cite{Li2023}. 

For completeness, we summarize the constitutive laws and plasticity projections adopted for the five material categories. 
Let $\mathbf{F}$ be the deformation gradient, $J = \det(\mathbf{F})$ its Jacobian, and $\mathbf{T}$ the Cauchy stress.
We also denote by $Z(\mathbf{F})$ a return-mapping operator that projects $\mathbf{F}$ back to the admissible
elastic region whenever a yield condition is violated.

\textbf{Elastic Solids}:
Elastic objects are modeled with the neo-Hookean law. 
The stress is written as:
\begin{equation}
  J\, \mathbf{T}(\mathbf{F}) 
  = 
  \mu \bigl(\mathbf{F}\mathbf{F}^\top\bigr) 
  + \bigl(\lambda \log J - \mu\bigr)\, \mathbf{I},
\end{equation}
where $\mu$ and $\lambda$ are Lam\'e parameters related to Young's modulus $E$ and Poisson
ratio $\nu$ via
\begin{equation}
  \mu = \frac{E}{2 (1 + \nu)}, 
  \qquad
  \lambda = \frac{\nu E}{(1 + \nu)(1 - 2\nu)}.
\end{equation}

\textbf{Plasticine}:
Plasticine is treated as an St.\ Venant--Kirchhoff (StVK) solid in Hencky (logarithmic) strain, combined with von-Mises plasticity. 
We write the singular value decomposition (SVD) of $\mathbf{F}$ as 
$\mathbf{F} = \mathbf{U}\boldsymbol{\Sigma}\mathbf{V}^\top$ and define
the Hencky strain $\boldsymbol{\epsilon} = \log \boldsymbol{\Sigma}$.
The elastic stress is
\begin{equation}
  J\, \mathbf{T}(\mathbf{F}) 
  =
  \mathbf{U}\bigl(2\mu\, \boldsymbol{\epsilon} 
  + \lambda\, \mathrm{tr}(\boldsymbol{\epsilon})\, \mathbf{I}\bigr)\mathbf{U}^\top.
\end{equation}
Let $\hat{\boldsymbol{\epsilon}}$ be the normalized Hencky strain and $\tau_Y$ the yield stress.
The von Mises yield measure is
\begin{equation}
  \delta \gamma 
  = 
  \|\hat{\boldsymbol{\epsilon}}\| 
  - 
  \frac{\tau_Y}{2\mu}.
\end{equation}
If $\delta\gamma \le 0$, the state is elastic and $Z(\mathbf{F}) = \mathbf{F}$.
Otherwise,
\begin{equation}
  Z(\mathbf{F}) 
  = 
  \mathbf{U}\exp\!\Bigl(
      \boldsymbol{\epsilon}
      - 
      \delta\gamma 
      \frac{\hat{\boldsymbol{\epsilon}}}{\|\hat{\boldsymbol{\epsilon}}\|}
    \Bigr)\mathbf{V}^\top .
\end{equation}

\textbf{Newtonian Fluids}:
For Newtonian fluids we use a $J$-based formulation with an additional viscous term. 
Let $\mathbf{v}$ denote the velocity field and $\nabla \mathbf{v}$ its spatial gradient.
The stress is defined as:
\begin{equation}
  J\, \mathbf{T}(\mathbf{F}) 
  = 
  \tfrac{1}{2}\,\mu \bigl(\nabla \mathbf{v} + \nabla \mathbf{v}^\top\bigr)
  + \kappa\bigl(J - J^{-6}\bigr).
\end{equation}

\textbf{Non-Newtonian Fluids}:
Non-Newtonian fluids are modeled with a viscoplastic extension of the above von Mises plasticity. 
Using the same SVD $\mathbf{F} = \mathbf{U}\boldsymbol{\Sigma}\mathbf{V}^\top$ and strain
$\boldsymbol{\epsilon}$, let $d$ be the spatial dimension and define
\begin{align}
\hat{\mu} &= \mu \,\frac{\mathrm{tr}(\boldsymbol{\Sigma}^2)}{d} \nonumber \\
\boldsymbol{s} &= 2\mu\, \hat{\epsilon} \\
\hat{s} &= \|\boldsymbol{s}\| - \frac{\delta\gamma}{1 + \tfrac{\eta}{2 \hat{\mu} \Delta t}} \nonumber
\end{align}
where $\eta$ is the plastic viscosity.  
The return map becomes
\begin{equation}
  Z(\mathbf{F})
  =
  \begin{cases}
    \mathbf{F}, & \delta\gamma \le 0,\\[3pt]
    \mathbf{U}\exp\!\Bigl(
       \tfrac{\hat{s}}{2\mu}\, \hat{\boldsymbol{\epsilon}}
       + 
       \tfrac{1}{d}\mathrm{tr}(\boldsymbol{\epsilon})\, \mathbf{1}
    \Bigr)\mathbf{V}^\top,
    & \text{otherwise},
  \end{cases}
\end{equation}
with $\eta = 0$ recovering the plasticine model.

\textbf{Granular Substances}:
Granular materials use a Drucker--Prager yield condition with the same StVK log-strain elasticity. With $\boldsymbol{\epsilon}$ and $\hat{\boldsymbol{\epsilon}}$ defined above, the yield criterion is
\begin{equation}
  \mathrm{tr}(\boldsymbol{\epsilon}) > 0
  \quad\text{or}\quad
  \delta\gamma
  =
  \|\hat{\boldsymbol{\epsilon}}\|_F
  +
  \alpha
  \frac{(d\lambda + 2\mu)\,\mathrm{tr}(\boldsymbol{\epsilon})}{2\mu}
  > 0 ,
\end{equation}
where $\alpha$ relates to the friction angle $\theta_{\mathrm{fric}}$ via
\begin{equation}
  \alpha 
  = 
  \sqrt{\tfrac{2}{3}}\,
  \frac{2\sin\theta_{\mathrm{fric}}}{3 - \sin\theta_{\mathrm{fric}}}.
\end{equation}
The return mapping is
\begin{equation}
  Z(\mathbf{F})
  =
  \begin{cases}
    \mathbf{U}\mathbf{V}^\top, & \mathrm{tr}(\boldsymbol{\epsilon}) > 0,\\[3pt]
    \mathbf{F}, & \delta\gamma \le 0,\ \mathrm{tr}(\boldsymbol{\epsilon}) \le 0,\\[3pt]
    \mathbf{U}\exp\!\Bigl(
       \boldsymbol{\epsilon}
       -
       \delta\gamma
       \tfrac{\hat{\boldsymbol{\epsilon}}}{\|\hat{\boldsymbol{\epsilon}}\|}
    \Bigr)\mathbf{V}^\top,
    & \text{otherwise}.
  \end{cases}
\end{equation}

Also, we integrate it into Blender 3.6.23 through an extended version of the Taichi Elements addon. The system supports the entire workflow from geometry import and discretization to GPU-parallel simulation, complex boundary handling, caching, and reconstruction for visualization.

Geometric assets are loaded directly inside Blender, including native meshes and BlenderKit models. For each object, we specify whether it behaves as a deformable body or a static collider and assign its material model, chosen from elastic solids, plasticine, Newtonian fluids, non-Newtonian fluids, or granular substances. The selected geometry is then discretized into an MPM representation using Taichi Elements: mesh surfaces are voxelized and sampled into particles, and all particle attributes (mass, velocity and material parameters, \etc{}.) are initialized and stored as Taichi GPU fields.

During simulation, each frame is advanced through several MPM substeps executed entirely on GPUs. In each substep, particle states are first transferred to the grid (P2G). For every particle, we compute a trial deformation update from its affine velocity field, apply the appropriate material-dependent deformation projection, evaluate the corresponding constitutive model, and transfer mass, momentum, and APIC affine terms to grid nodes through quadratic interpolation. This stage fully determines the behavior of different materials and follows the exact constitutive formulations of PAC-NeRF. After P2G, grid velocities are normalized, external forces such as gravity are applied in a dedicated kernel.

Boundary interactions are handled in a subsequent grid post-processing phase, where several collider kernels are executed sequentially. These include a global axis-aligned bounding box, analytic primitive colliders such as planes and spheres, and a complex mesh collider capable of handling arbitrary Blender geometry. Colliders operate directly on grid velocities by projecting or clamping them according to the selected contact mode (sticky, slip, or separate). The updated grid velocities are then transferred back to particles (G2P), after which particle positions, velocities, and affine fields are advanced.

To robustly support collisions with highly detailed Blender meshes, we implement a voxelized Signed Distance Field (SDF) collider. Input meshes are voxelized into a regular grid, inside/outside regions are identified, and each voxel is assigned a signed distance value and an estimated normal. During simulation, grid nodes near the mesh surface query the SDF and adjust their velocity according to the relevant contact rules. This approach provides a stable, high-resolution, and geometry-agnostic collision handling mechanism that integrates seamlessly with the Taichi Elements pipeline. To enhance dataset diversity, we stochastically vary the simulation parameters (\eg{}, initial pose, velocity, or material coefficients) during refinement. In the Taichi Elements addon, parameters such as the simulation domain, simulation resolution, and time step are set for the scene. We use a uniform parameter configuration: $dt=1/150, dx=dy=dz=0.0083$.

In the visualization pipeline, continuous surfaces are reconstructed from particles using Blender Geometry Nodes, with standard Blender materials and rendering engines applied to enhance realism. However, for granular substances, we opt for direct particle rendering to accurately capture their discrete nature. Regarding material representation, commonly referenced physical materials are used as benchmarks: for instance, honey serves as a uniform rendering material for viscous Newtonian fluids, while sand is adopted for granular substances. The whole procedure yields 240 scenes per material category, resulting in a total of 1,200 scenes.

\subsubsection{Physical Parameter Range}\label{sec:physical parameter range}
We report the ranges of example physical parameters in Table~\ref{tab:phys_params}. The wide ranges of these parameters highlights the diversity of our dataset.
\begin{table}[hb]\tabcolsep= 0.06cm  \vspace{-0.35cm} 
\centering
\caption{Ranges of Example Physical Parameters.}
\label{tab:phys_params}\vspace{-0.4cm}
\resizebox{0.48\textwidth}{!}{
\begin{tabular}{lcccccccccc}
\toprule
& \begin{tabular}[c]{@{}c@{}}Friction \\ Coefficient\end{tabular} &  Restitution & Density & \begin{tabular}[c]{@{}c@{}}SPH \\ Viscosity\end{tabular}   & \begin{tabular}[c]{@{}c@{}}SPH Surface \\ Tension\end{tabular} & \begin{tabular}[c]{@{}c@{}}Fluid\\Viscosity\end{tabular} &  \begin{tabular}[c]{@{}c@{}}Plasticity\\Viscosity\end{tabular} & \begin{tabular}[c]{@{}c@{}}Yield\\Stress\end{tabular}  & \begin{tabular}[c]{@{}c@{}}Friction\\Angle\end{tabular} & ..  \\
\midrule
min $\sim$ max & 0. $\sim$ 0.9  & 0.1 $\sim$ 0.8 & 0.1 $\sim$ 21.3 & 0.0 $\sim$ 5.0 & 0.5 $\sim$ 3.0 & 25 $\sim$ 200 & 3 $\sim$ 10 &  0.1 $\sim$ 10 & 15 $\sim$ 60 & .. \\
\bottomrule
\end{tabular}
}\vspace{-0.4cm}
\end{table}

\subsection{More Details of Cameras and Rendering}\label{app:cameras_render}

\subsubsection{Static Camera Position Sampling}
\begin{figure*}[t!]
\centering
\includegraphics[width=0.9\linewidth]{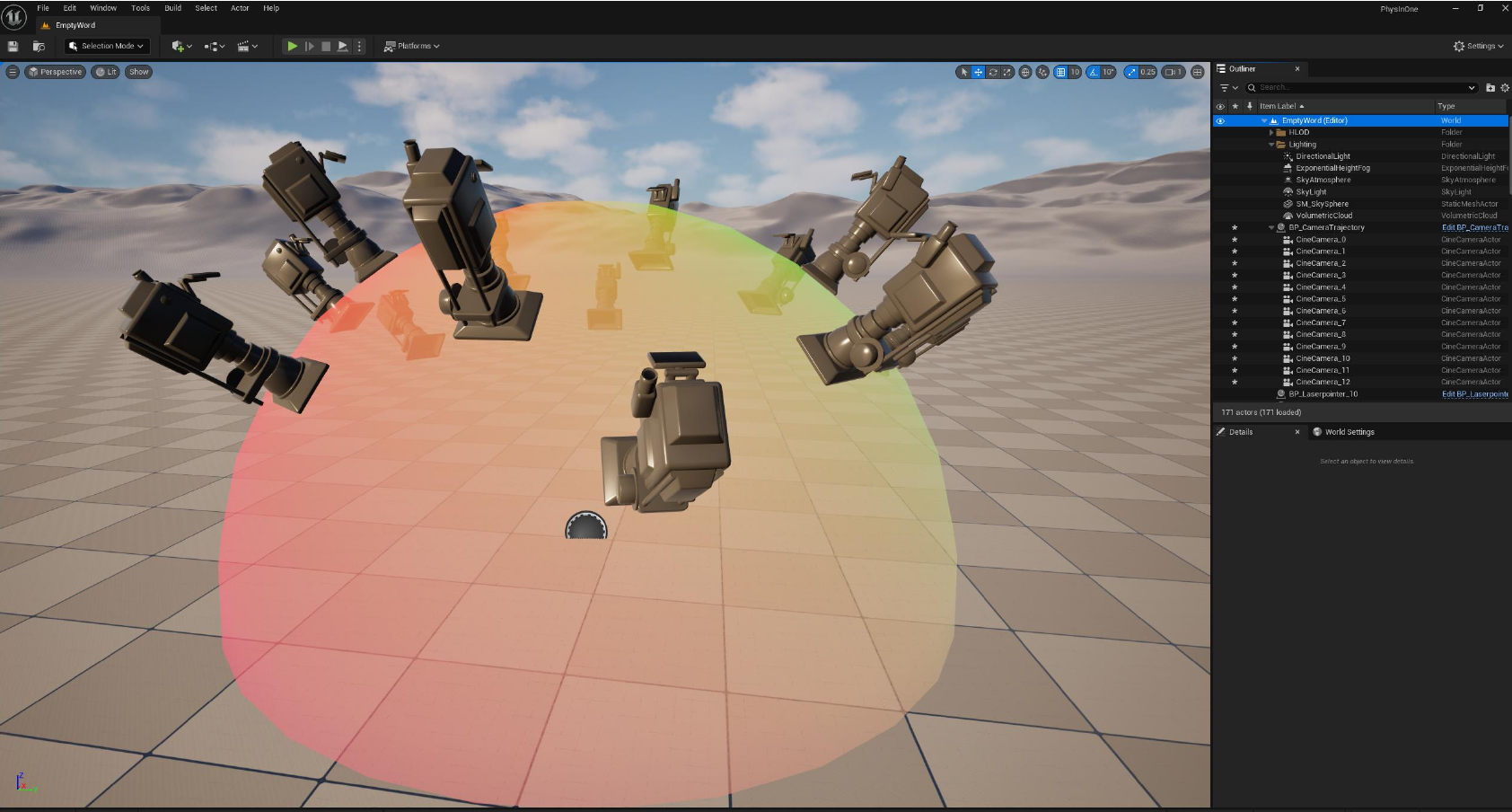}
\vskip -0.1in
\caption{Static Camera Sampling}
\label{fig:static_camera}
\vspace{-0.2cm}
%\vskip 0.4in
\end{figure*}

The cameras are positioned on a spherical surface whose center coincides with the physical event's origin. The sphere's radius is chosen such that it fully encompasses all objects involved in the event as well as their expected motion range, ensuring complete coverage of the phenomenon. Initially, we uniformly sample 12 static cameras on the upper hemisphere, with camera viewpoints evenly distributed within a latitude range of $30^\circ$ to $60^\circ$. Cameras that suffer from occlusions are resampled, ensuring that all 12 cameras are well positioned and provide unobstructed coverage. Figure~\ref{fig:static_camera} shows an example of static camera distribution.

For special material and liquid simulations, we leverage Blender to generate cameras and render corresponding RGB images, depth and segmentation information. The static camera viewpoints are uniformly sampled from the upper hemisphere, 15 camera viewpoints for special material simulations and 12 camera viewpoints for liquid simulations. Dynamic camera trajectories are generated based on predefined camera sampling strategies.

\subsubsection{Monocular Camera Trajectory Sampling}
To render monocular videos, we design three trajectory sampling strategies as detailed in below, and one of them is randomly selected for each dynamic 3D scene. Each strategy returns $N$ sampled positions, which are used to interpolate more positions, forming a smooth trajectory whose temporal resolution matches the number of frames in the video.

\begin{figure*}[t!]
\centering
\includegraphics[width=0.9\linewidth]{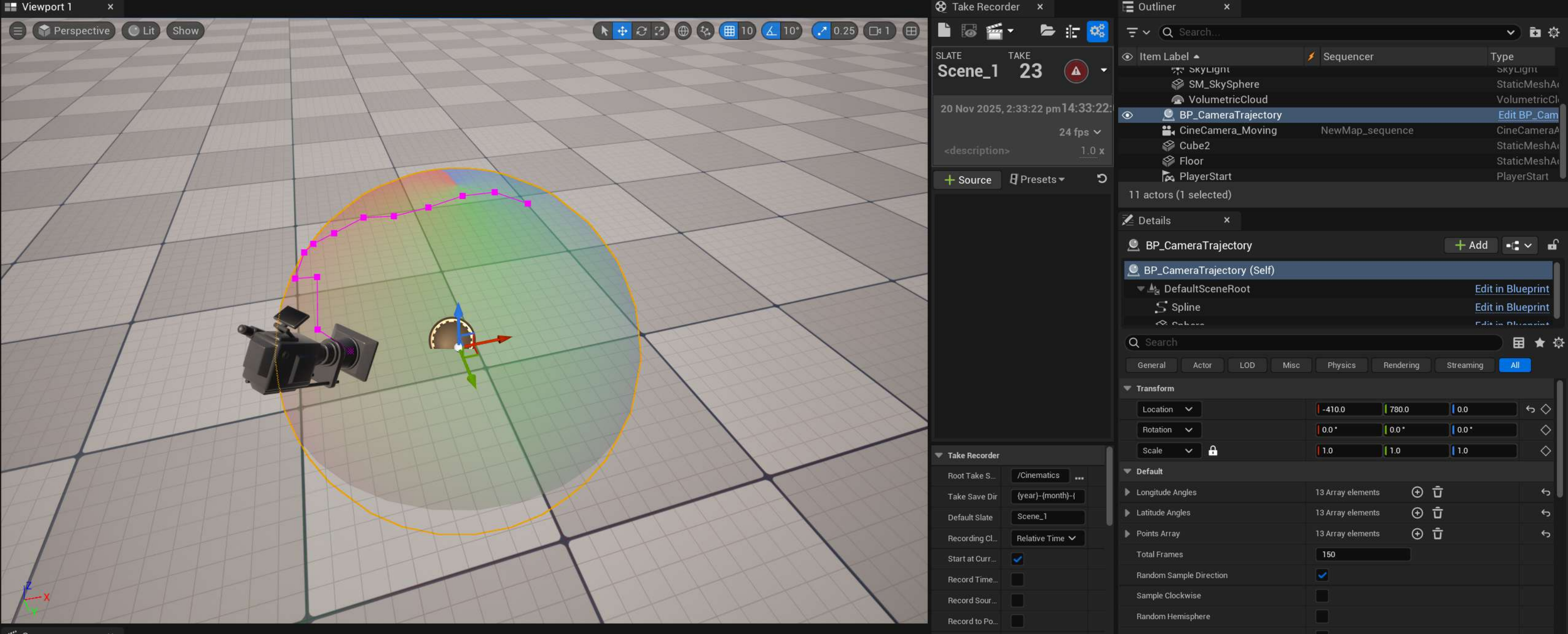}
\vskip -0.1in
\caption{Linear Drift Sampling}
\label{fig:trajectory_1}
\vspace{-0.2cm}
%\vskip 0.4in
\end{figure*}

\textbf{Linear Drift Sampling}:
This method generates a trajectory by evenly spacing sample positions along longitude within a fixed range of $180^\circ$, starting from a randomly chosen initial longitude. The direction of progression (clockwise or counterclockwise) is selected randomly. For latitude, the first position is sampled within the range of $[-45^\circ, 45^\circ]$, respecting hemisphere constraints (upper, lower, or both). The number of sample points is randomly chosen between 10 and 20. Subsequent latitude values are obtained by adding random perturbations to the previous point, ensuring each new latitude remains within the allowed range and satisfies hemisphere restrictions. If a valid latitude cannot be found after repeated attempts, a fallback value is used. This approach creates a trajectory that is globally structured along longitude but locally varied in latitude, balancing deterministic progression with stochastic variation. Figure~\ref{fig:trajectory_1} shows an example of camera trajectory.

\begin{figure*}[t!]
\centering
\includegraphics[width=0.9\linewidth]{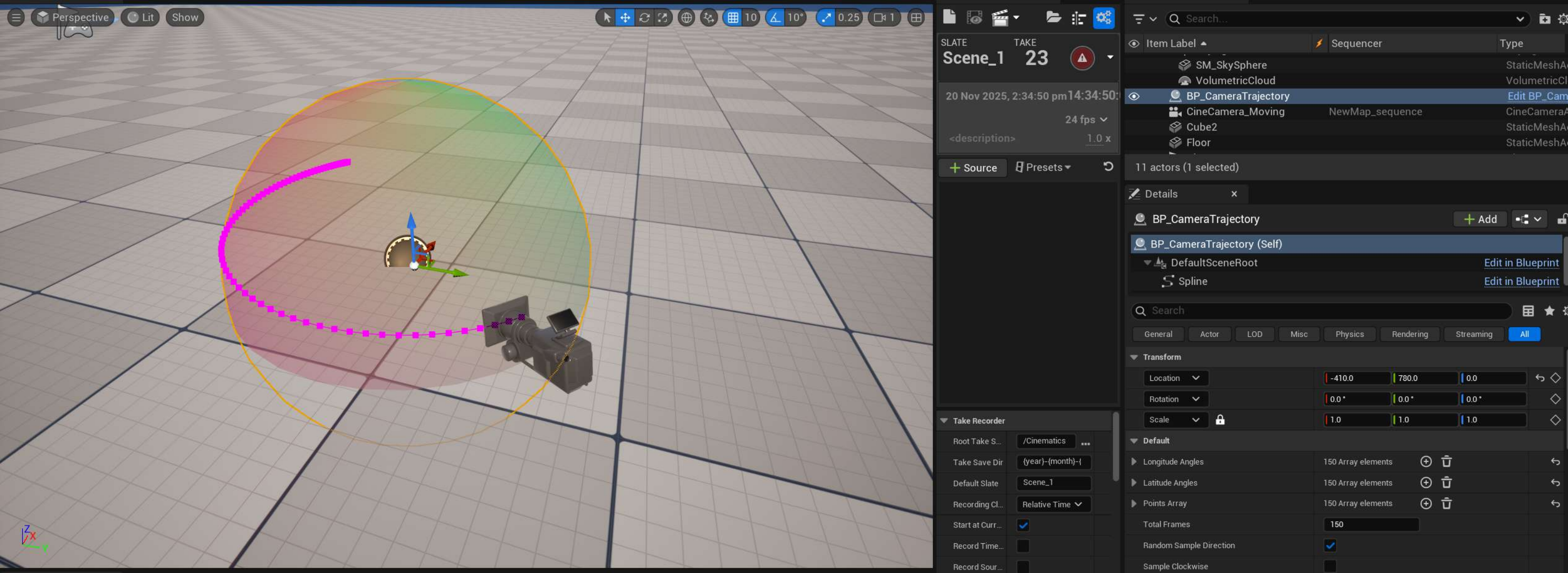}
\vskip -0.1in
\caption{Sinusoidal Interpolation Sampling}
\label{fig:trajectory_2}
\vspace{-0.2cm}
%\vskip 0.4in
\end{figure*}

\textbf{Sinusoidal Interpolation Sampling}:
This method generates a smooth trajectory between a randomly sampled start point and end point on the sphere. Both points are sampled within a longitude span of $180^\circ$ and a latitude range of $[-45^\circ, 45^\circ]$, respecting hemisphere constraints (upper, lower, or both). The trajectory is interpolated using sinusoidal functions, ensuring a gradual transition in both longitude and latitude rather than abrupt changes. A dynamic radius factor is applied along the path to simulate natural camera movement. Compared to Linear Drift Sampling, this approach produces a more structured and aesthetically pleasing curve, generating smooth viewpoint transitions. Figure~\ref{fig:trajectory_2} shows an example of camera trajectory.

\begin{figure*}[ht!]
\centering
\includegraphics[width=0.9\linewidth]{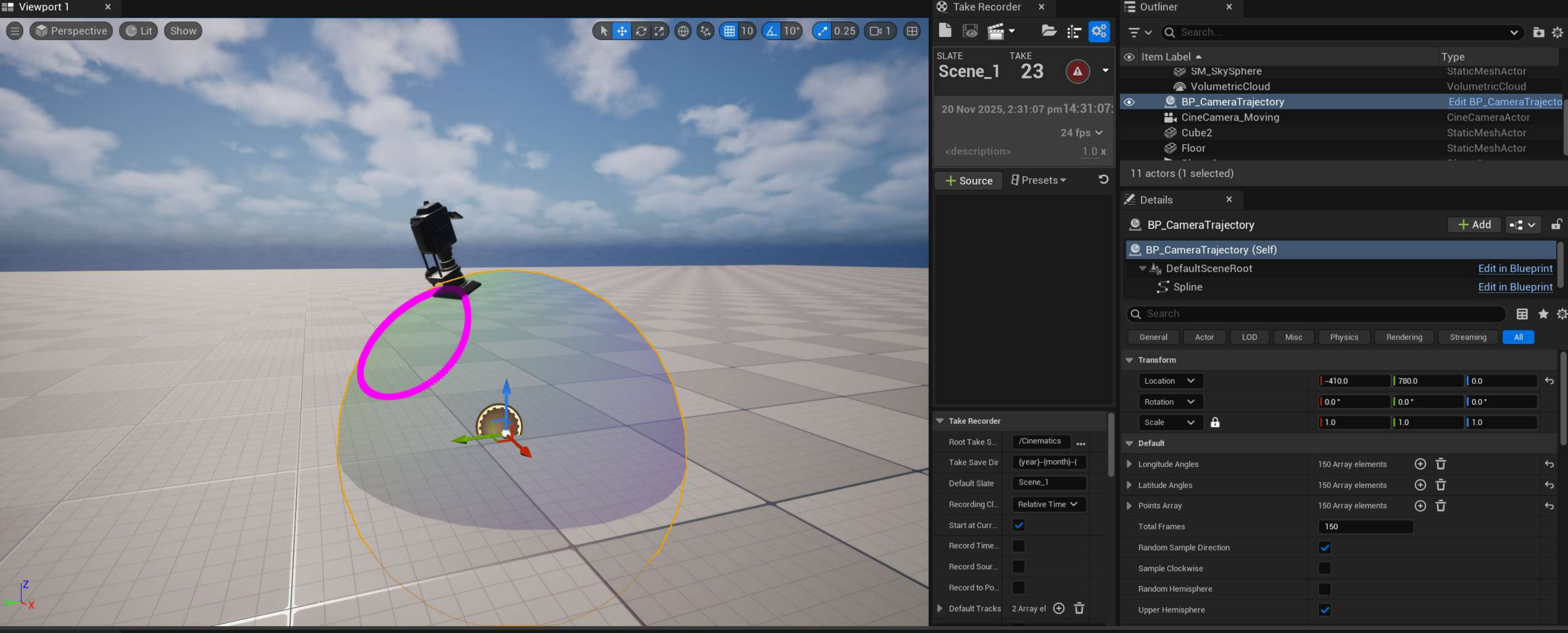}
\vskip -0.1in
\caption{Circular Loop Trajectory Sampling on Sphere}
\label{fig:trajectory_3}
\vspace{-0.2cm}
%\vskip 0.4in
\end{figure*}

\textbf{Circular Loop Trajectory Sampling on Sphere}:
This method samples points along a circular trajectory on a sphere.
A loop center is randomly chosen within hemisphere constraints, and a circle of adjustable size is defined by a loop intensity parameter.
The trajectory is parameterized by $n$ evenly spaced angles, generating latitude and longitude offsets that form a closed loop around the center.
Hemisphere clipping ensures valid positions, and an optional radial variation introduces diversity.
This approach provides structured yet flexible sampling for spherical domains.
Figure~\ref{fig:trajectory_3} shows an example of a camera trajectory.

\begin{figure}[ht!]
\centering
\includegraphics[width=1\linewidth]{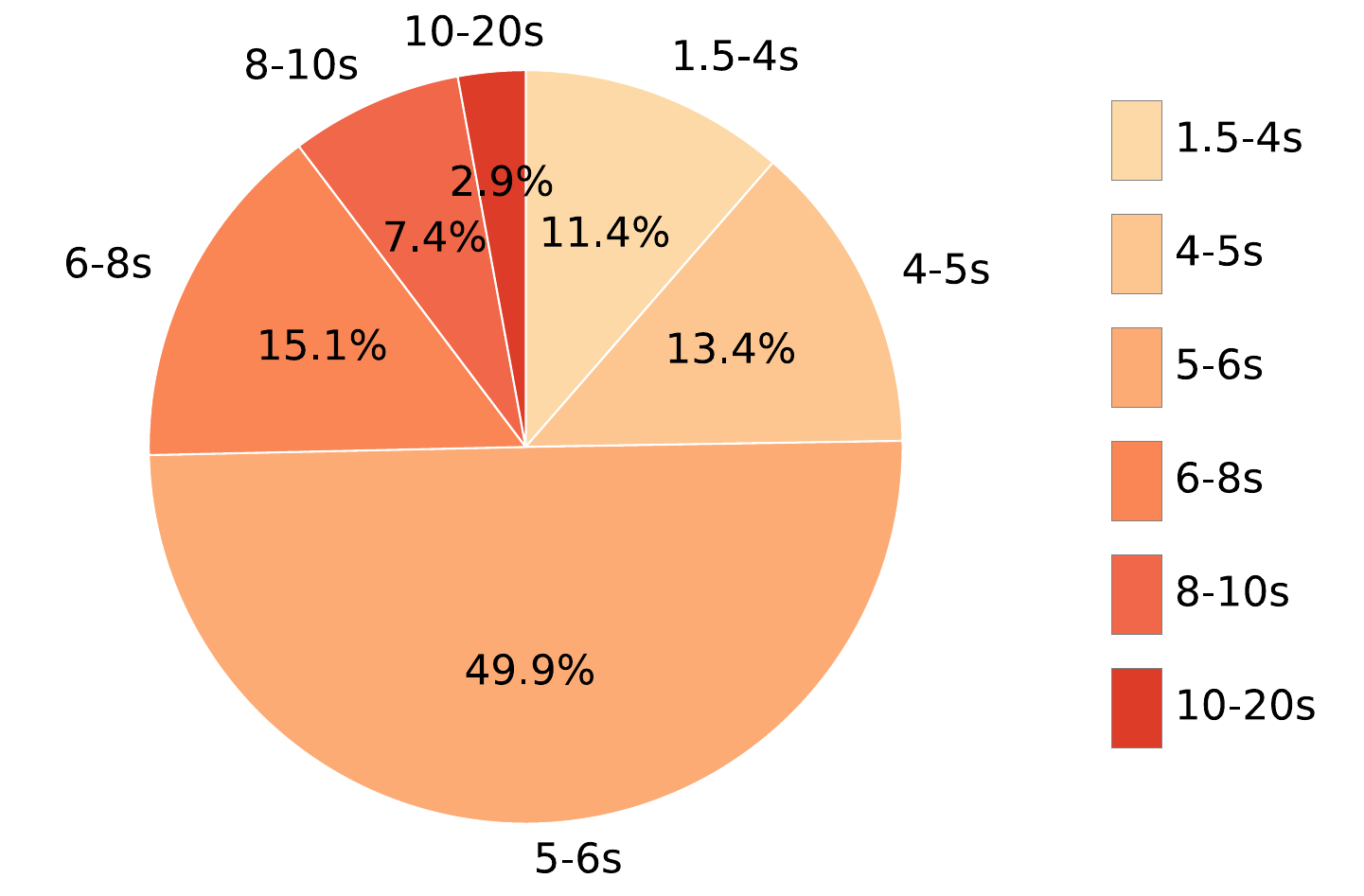}
\vskip -0.1in
\caption{The distribution of video lengths.}
\label{fig:duration_distribution}
\vspace{-0.4cm}
%\vskip 0.4in
\end{figure}

\subsubsection {Distribution of Video Lengths}
Our dataset consists of 2 million videos generated from 153,810 unique dynamic 3D scenes. Figure \ref{fig:duration_distribution} presents the distribution of video lengths within the dataset, highlighting that a substantial proportion, approximately 75\%, of the videos are long-horizon, lasting longer than 5 seconds ($\geq 150$ frames). This characteristic makes our dataset particularly well-suited for a variety of challenging tasks in long-horizon, physically-grounded video generation, simulation,  understanding, \etc{}.

\subsection{More Details of Annotating and Splitting Data}\label{app:annotate_split}

\subsubsection{Annotating}
Our dataset has annotations of geometry, semantics, motion, physical properties, and textual descriptions.

\begin{itemize}
    \item \textbf{Geometry}: Depth maps measured in meters with a resolution of $1120 \times 1120$ are rendered with RGB images. Meshes of all 3D assets are also available. 
    \item \textbf{Semantics}: Segmentation maps with a resolution of $1120 \times 1120$, where each foreground object is assigned a unique ID starting from 1 and the background is assigned 0, are also rendered with RGB images.
    \item \textbf{Motion}: Each object is associated with a local coordinate origin. By recording the position and rotation of this origin in the world coordinate system at every time step, we capture the object's 3D trajectory. The 3D trajectory of any point on the object can be reconstructed using its local coordinates. For rigid objects, based on the camera information, depth and segmentation maps, we can reconstruct the 3D points. Through the position and rotation at each time step, we can calculate the transformation of the object between different frames, and thus we can obtain the dense flow for rigid objects. 
    \item \textbf{Physical Properties}: Each object is linked to a physical material characterized by four key parameters: dynamic friction coefficient, static friction coefficient, density, and restitution coefficient. These parameters are stored in a JSON format, where the key corresponds to the object's segmentation ID and the value contains its physical attributes.
    \item \textbf{Text Annotating}: The text annotations are manually curated to inject human-level understanding of dynamics and physics into our dataset, addressing a key limitation of current state-of-the-art video understanding pipelines, which often fail to explain scenes involving highly coupled objects and complex, varying dynamics, let alone perform underlying physical reasoning. Specifically, we recruit human annotators to watch each scene at least three times from at least three different camera viewpoints. For each scene, they are instructed to describe:  
\begin{enumerate}
\item The initial state of every object (\eg{}, position, orientation, initial velocity);
\item The dynamic behavior of each object together with the governing physical principles (\eg{}, \textit{``the white football falls due to gravity and rebounds upon collision with the ground''});
\item The dominant physical laws illustrated by the scene as a whole (\eg{}, conservation of momentum, elastic collision).
\end{enumerate}

With accurate human annotations, we refine the descriptions using Qwen3-VL-235B-A22B-Thinking, a large language model, to correct grammatical errors, improve clarity, and enhance completeness. During this polishing step, we provide an additional prompt that emphasizes object appearance details and explicitly highlights the relevant physics laws, ensuring the final annotations are both fluent and physically informative.
\end{itemize}

\subsubsection{Dataset Splitting}
Each scene contains 1 $\sim$ 3 physical phenomena from Table~\ref{tab:list-physics-phen-law}. We partition all scenes into train, validation, and test splits at the ratio of 8:1:1, while ensuring that the proportion of each physical phenomenon is approximately preserved across splits. In the meantime, 3D assets in different partitions appear exclusively. 

\subsubsection{Dataset Quality}
Our dataset quality is ensured through a structured workflow, including standardized assets, seasoned UE developers, multiple rounds of data validation by independent reviewers using unified criteria.

\subsection{More Details of Video Generation}\label{app:vid_gen_exp}

\subsubsection{Experiment Setup}
We evaluate three representative video generation models on our dataset: Stable Video Diffusion XL (\textbf{SVD})\cite{Blattmann2023}, an image-to-video model, and two image-text-to-video models: CogVideoX-1.5-5B (\textbf{CogVideoX})\cite{Yang2025} and Wan 2.2-5B (\textbf{Wan2.2})\cite{Wan2025}.

SVD is based on a U-Net architecture, where the diffusion denoiser follows a U-shaped design. It conditions the generation process on the first-frame image by concatenating input noise with the image and computing cross-attention between the denoising hidden states and image features extracted from a pre-trained CLIP encoder at multiple layers.
In contrast, both Wan2.2 and CogVideoX adopt a Diffusion Transformer (DiT) architecture. Their key difference lies in how they incorporate conditional information: Wan2.2 injects the text prompt via cross-attention layers alongside image conditioning, whereas CogVideoX concatenates all condition signals, including the reference image and text embeddings, into a single long token sequence that is jointly processed by the transformer.

In this paper, we fine-tune each model until convergence on a subset of our dataset, with about 1K training steps for SVD, 2K steps for Wan2.2, and 1K steps for CogVideoX. To provide practical guidance for future users interested in adapting these models to our data, we explore three distinct fine-tuning strategies: Low-Rank Adaptation (LoRA), Supervised Fine-Tuning (SFT), and Final-Layer Tuning (FLT), resulting in a suite of adapted variants for each architecture.

All experiments use a learning rate of 0.0001 and a batch size of 64. For LoRA, we set the rank to 32 and the scaling factor $\alpha$ to be 1.0 uniformly across all applicable layers. For FLT, we freeze all parameters except those in the final two transformer blocks (or equivalent decoder layers), yielding approximately 400 MB of trainable parameters per model. 

\subsubsection{Details of PMF}

The key motivation of introducing Fourier based metrics for evaluating generated videos is that we aim to focus on evaluating \textbf{motion dynamics} rather than the absolute correctness of pixel values. Specifically, given an AI-generated video $\mathcal{V}_{\text{gen}}$ and a reference video $\mathcal{V}_{\text{ref}}$, we apply the 3D Discrete Fourier Transform (DFT) independently to each video across its spatial and temporal dimensions ($H \times W \times T$):

\begin{equation}
    \mathcal{\tilde{V}}(c; u, v, s) = \sum_{h, w, t} \mathcal{V}(c; h, w, t) \cdot e^{-2\pi i \left( \frac{u h}{H} + \frac{v w}{W} + \frac{s t}{T} \right)}
\end{equation}
where $\tilde{\mathcal{V}}(c; u, v, s)$ denotes the Fourier coefficients at spatial-temporal frequency indices $(u, v, s)$ for color channel $c \in \{0,1,2\}$ (corresponding to RGB), capturing spatial frequencies $(u, v)$ and temporal frequency $s$. For brevity, we use $\sum_{h,w,t}$ as shorthand for the full spatiotemporal summation $\sum_{h=0}^{H-1} \sum_{w=0}^{W-1} \sum_{t=0}^{T-1}$, which aggregates over all spatial and temporal positions or frequencies in the video.

The following well-known properties of the Fourier transform are instrumental to our analysis:

\medskip

We further define the amplitude $A_{c; u, v, s}$, phase $\psi_{c; u, v, s}$, and normalized energy $E_{u, v, s}$ for each frequency component per color channel as follows:

\begin{equation}
    A_{c; u, v, s} = \big| \tilde{\mathcal{V}}(c; u, v, s) \big|
\end{equation}

\begin{equation}
\begin{aligned}
   \psi_{c; u, v, s} &= \arg\big( \tilde{\mathcal{V}}(c; \vec{\omega}) \big) \\
    &= \arctan\!\left( \frac{\operatorname{Im}\big( \tilde{\mathcal{V}}(c; u, v, s) \big)}{\operatorname{Re}\big( \tilde{\mathcal{V}}(c; u, v, s) \big)} 
    \right)
    \label{eq:fft_phase_definition}
\end{aligned}
\end{equation}

\begin{equation}
    E_{u, v, s} = \frac{ \sum_{c=0}^{2} \big| \tilde{\mathcal{V}}(c;  u, v, s) \big|^2 }{  \sum_{u',v',s'}  \sum_{c=0}^{2} \big| \tilde{\mathcal{V}}(c; u', v', s') \big|^2 }
    \label{eq:fft_energy_definition}
\end{equation}

Crucially, under the standard assumption that the video signal is extended beyond its spatial and temporal boundaries via duplicate padding, we consider (1) a spatiotemporal shift, denoted by $\delta \mathbf{x} = (\delta h, \delta w, \delta t)$, and (2) a brightness rescaling in the original video signal, denoted by $\lambda \in (0, 1]$.
This yields a shifted video $\mathcal{V}'$ defined as:
\begin{equation}
    \mathcal{V}'(c; h, w, t) = \lambda\mathcal{V}(c; h + \delta h, w + \delta w, t + \delta t)
\end{equation}

Then we apply Fourier transform on $\mathcal{V}'$:

\begin{equation}
\begin{aligned}
& \tilde{\mathcal{V}}'(c; u, v, s) \\
&=  \sum_{h,w,t}
\lambda \mathcal{V}(c; h + \delta h, w + \delta w, t + \delta t) \,
e^{-2\pi i \left( \frac{u h}{H} + \frac{v w}{W} + \frac{s t}{T} \right)} \\
&=  \sum_{h',w',t'} 
\lambda \mathcal{V}(c; h', w', t') \,
e^{-2\pi i \left( \frac{u (h' - \delta h)}{H} + \frac{v (w' - \delta w)}{W} + \frac{s (t' - \delta t)}{T} \right)} \\
&= \lambda e^{i \delta\psi}\sum_{h',w',t'}\mathcal{V}(c; h', w', t') \,
e^{-2\pi i \left( \frac{u h'}{H} + \frac{v w'}{W} + \frac{s t'}{T} \right)} \\
&= \lambda e^{i \delta\psi} \tilde{\mathcal{V}}(c; u, v, s)
\end{aligned}
\end{equation}
where $\delta\psi = 2\pi\left( \frac{u \delta h}{H} + \frac{v \delta w}{W} + \frac{s \delta t}{T} \right)$.
Note that we assume duplicate padding to prevent boundary artifacts induced by the shift, thereby allowing the summation limits to be extended validly.

\begin{figure*}[ht!]
\centering
\includegraphics[width=1.\linewidth]{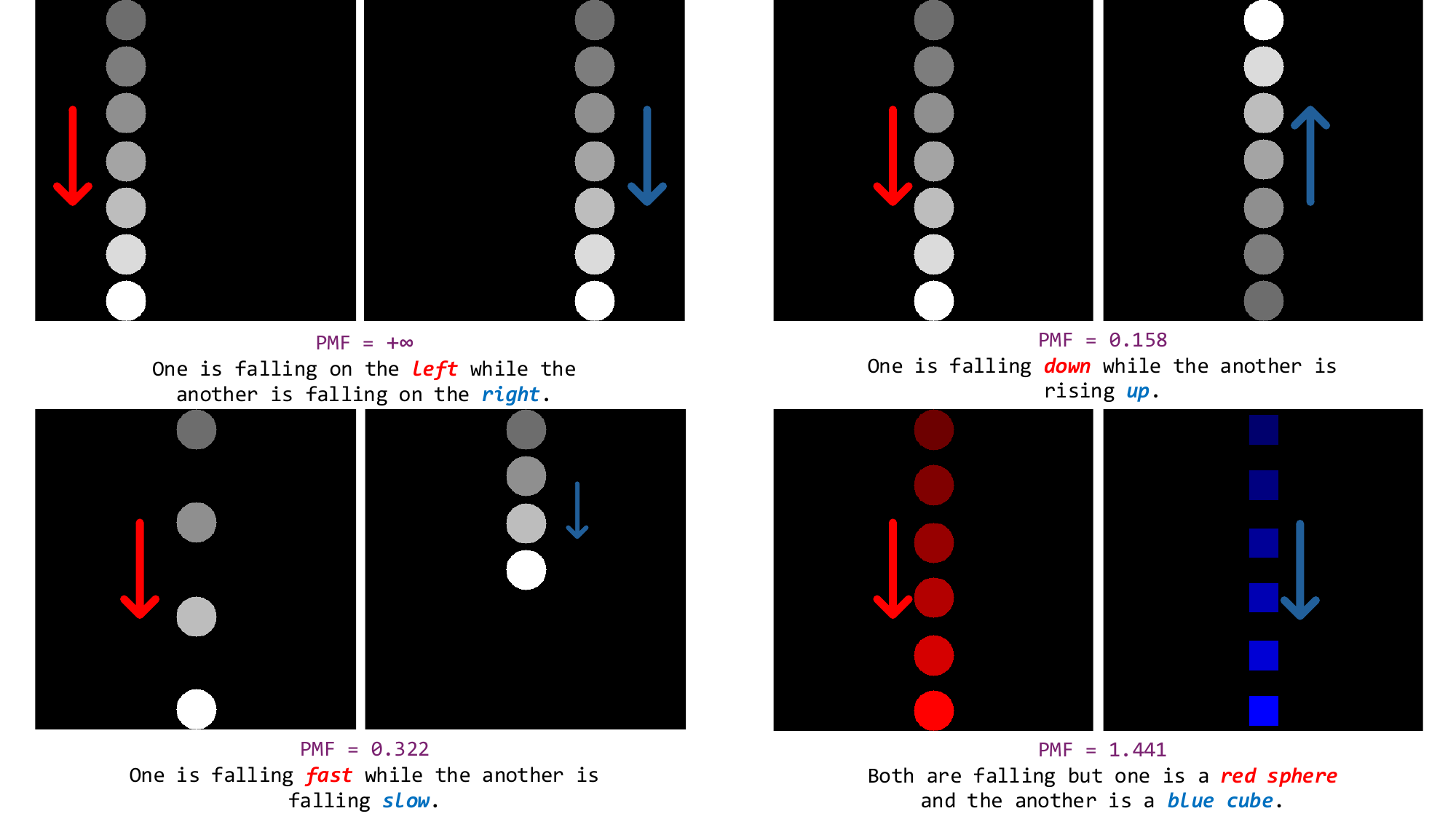}
\vskip -0.1in
\caption{Demonstration for PMF. In the top-left pair, the only variance is the initial spatial location where the object begins to fall. As expected, \text{PMF} yields a perfect score, reflecting identical motion patterns regardless of absolute positioning.
Conversely, in the top-right pair, the motion is completely reversed (\eg{}, upward vs. downward fall). The metric degrades severely due to the starkly mismatched spectral energy distributions, correctly indicating poor motion similarity.
In the bottom-left case, the initial position is fixed but the velocity is halved. Here, the \text{PMF} value increases substantially, effectively distinguishing the altered temporal dynamics through the energy spectrum.
Finally, in the bottom-right example, where the object shape differs but the underlying motion dynamics are preserved, the \text{PMF} remains relatively high. This confirms that \text{PMF} is largely invariant to visual appearance variations and focuses primarily on the consistency of motion dynamics.}
\label{fig:fourier_toy}
\vspace{-0.4cm}
%\vskip 0.4in
\end{figure*}

Consequently, the magnitude, normalized energy, and phase transform as:

\begin{equation}
\begin{aligned}
A'_{c; u,v,s} &= \big| \tilde{\mathcal{V}}'(c; u, v, s) \big| \\
&= \big| \lambda e^{i \delta\psi}\tilde{\mathcal{V}}(c; u, v, s) \big| \\
&= \lambda \big| \tilde{\mathcal{V}}(c; u, v, s) \big| \\
&= \lambda A_{c; u,v,s} \\
\end{aligned}
\end{equation}

\begin{equation}
\begin{aligned}
E'_{u,v,s} 
&= \frac{ \sum_{c=0}^{2} \big| \tilde{\mathcal{V}}'(c; u, v, s) \big|^2}
        { \sum_{u',v',s'} \sum_{c=0}^{2} \big| \tilde{\mathcal{V}}'(c; u', v', s') \big|^2} \\[6pt]
&= \frac{ \sum_{c=0}^{2} \big| \lambda \, \tilde{\mathcal{V}}(c; u, v, s) \big|^2}
        { \sum_{u',v',s'} \sum_{c=0}^{2} \big| \lambda \, \tilde{\mathcal{V}}(c; u', v', s') \big|^2} \\[6pt]
&= \frac{\lambda^2  \sum_{c=0}^{2} \big| \tilde{\mathcal{V}}(c; u, v, s) \big|^2}
        {\lambda^2  \sum_{u',v',s'} \sum_{c=0}^{2} \big| \tilde{\mathcal{V}}(c; u', v', s') \big|^2} \\[6pt]
&= E_{u,v,s}.
\end{aligned}
\end{equation}

\begin{equation}
\begin{aligned}
\psi'_{c; u,v,s} &= \psi_{c; u,v,s} + \delta\psi  \\
&=  \psi_{c; u,v,s} + 2\pi \left( \frac{u \delta h}{H} + \frac{v \delta w}{W} + \frac{s \delta t}{T} \right)
\end{aligned}
\end{equation}

From the above derivation, we can see that, a spatiotemporal shift affects only the phase of Fourier coefficients, leaving the amplitude (and hence the energy $E_{u,v,s}$) invariant, while brightness rescaling does not affect the normalized energy or the phase. In a nutshell, the normalized energy is invariant under any spatiotemporal shift and brightness rescaling, and the phase is invariant under brightness rescaling but encodes the information of the spatiotemporal translation.

Hence, we observe that the motion dynamics in $\mathcal{V}'$ are identical to those in $\mathcal{V}$, that only the initial spatiotemporal state (\ie{}, \textit{when} or \textit{where} the motion begins) and the absolute brightness differ. 
This insight inspires us to evaluate motion dynamics by comparing the differences between energy spectra.

We propose \textbf{PMF (Physical
Motion Fidelity)}, a metric that compares the similarity between the normalized spectral energy distributions of $\mathcal{V}_{\text{gen}}$ and $\mathcal{V}_{\text{ref}}$ as follows:
\begin{equation}
    \text{PMF}(\mathcal{V}_{\text{gen}}, \mathcal{V}_{\text{ref}}) = 
    -\ln d_{TV}(E^{\text{gen}}, E^{\text{ref}})
\end{equation}

where $d_{TV}$ refers to total variation that quantifies the difference between this two distributions:
\begin{equation}
    d_{TV}(E^{\text{gen}}, E^{\text{ref}}) = 
    \frac{1}{2} \sum_{u,v,s} 
    \left| E_{u,v,s}^{\text{gen}} - E_{u,v,s}^{\text{ref}} \right|
\end{equation}

where $E_{u,v,s}^{\text{gen}}$ and $E_{u,v,s}^{\text{ref}}$ denote the normalized energy spectra (as defined in Equation~\ref{eq:fft_energy_definition}) of the generated and reference videos, respectively.

Since the energy spectrum $E_{u,v,s}$ is invariant to spatiotemporal shifts and brightness rescaling, \text{PMF} effectively discards misalignment in the initial state and focuses solely on the similarity of motion dynamics across spatial and temporal dimensions. 
A higher \text{PMF} score indicates greater similarity in motion patterns between $\mathcal{V}_{\text{gen}}$ and $\mathcal{V}_{\text{ref}}$. 
By comparing the energy spectra of $\widetilde{\mathcal{V}}_{\text{gen}}$ and $\widetilde{\mathcal{V}}_{\text{ref}}$, we quantify how well the generated video preserves natural motion patterns in the reference video, thereby decoupling differences in the initial spatiotemporal state.

To demonstrate the distinct focuses of \text{PMF}, we present several toy examples in Figure~\ref{fig:fourier_toy}.
We can see that \text{PMF} is sensitive to differences in motion patterns but invariant to simple spatiotemporal shifts or color changing, yielding high scores when dynamics match regardless of starting position. Meanwhile, it degrades when motion is fundamentally altered.

\subsubsection{Details of Human Rating}
To assess the physical plausibility of generated videos from the perspective of human commonsense, we conduct a human evaluation study on 801 AI-generated videos. 

We recruit 34 human evaluators to rank the videos within each group according to heuristics that reflect commonsense physical reasoning. 
Before evaluating any generated videos, each participant is first shown the corresponding reference video to establish a ground-truth baseline for expected motion and object behavior. 
The heuristics are as follows: (1) \textbf{Trajectory Fidelity}: dynamic objects should follow motion paths that closely match the reference video in timing, direction, acceleration, and interaction; (2) \textbf{Prompt-Aligned Physics}: 
the dynamic objects should respect the physical constraints implied by the text prompt and remain generally plausible; (3) \textbf{Trajectory Smoothness}: the dynamic object motion should be temporally coherent and free of abrupt, jittery, or kinematically impossible transitions; (4) \textbf{Shape Consistency}: object geometry should remain stable over time, without unnatural deformations or flickering; (5) \textbf{Semantic Consistency}: object identity and category should be preserved (\eg{}, a ball should not transform into a bird); and (6) \textbf{Object Persistence}: no objects should spontaneously appear, disappear, or duplicate, namely basic conservation of existence must hold. 

Videos exhibiting severe physics violations beyond these categories (\eg{}, unexplained gravity reversal, teleportation, or momentum non-conservation) are ranked lowest unless explicitly justified by the prompt. Human evaluators are asked to provide a strict total ordering within each group (no ties), with the origins of AI models concealed and the presentation order of the generated videos fully randomized to minimize bias. Rankings are converted to scores, where the best rank corresponds to the highest score. 
The final human evaluation scores for each model are averaged across all groups and participants. Higher scores indicate better physical plausibility as judged by human assessors.

\subsection{More Details of Future Frame Prediction}\label{app:future_pred_exp}

\subsubsection{Details of Long-term Prediction}

We randomly select 103 dynamic 3D scenes from the test set. Each scene contains 10 seen viewing angles and 2 novel viewing angles. For each scene, we evaluate the model's extrapolation performance on both seen and novel views using four metrics: PMF, PSNR, SSIM, LPIPS.

For frame-based metrics (PSNR, SSIM, LPIPS), scores are averaged across views and time steps.
For PMF, scores are computed based on the predicted future video clip.

For FreeGave\cite{Li2025c}, we use videos from 10 viewing angles in training, utilizing the first half of each video's length. 
Initialization is performed by obtaining depth maps and camera poses for the first frame of all 10 views, back-projecting to generate approximately $1120 \times 1120 \times 10$ Gaussians, and randomly sampling 100{,}000 Gaussians as initialization. The model is optimized for 25{,}000 iterations. During testing, future frame prediction of the second half of video length is evaluated both on the 10 seen viewpoints and 2 novel viewpoints. DefGS\cite{Yang2024c}, TRACE\cite{Li2025b} and TiNeuVox\cite{Fang2022} use the same training and testing settings as FreeGave.

ExtDM\cite{Zhang2024} uses 83{,}650  videos for training. Stage 1 trains an autoencoder with settings: \texttt{max\_frame\_distance = 20}, \texttt{batch size = 256}, \texttt{image size = 64$\times$64}, \texttt{lr = 2e-4}, for 500{,}000 iterations. Each iteration samples two frames from a random video with frame distance $\leq$ \texttt{max\_frame\_distance}. Stage 2 trains a diffusion model with settings: \texttt{batch size = 64}, \texttt{image size = 64$\times$64}, \texttt{lr = 2e-4}, \texttt{condition frames = 30}, \texttt{pred frames = 20}, for 20{,}000 iterations. Each iteration samples 50 frames: the first 30 as condition frames and the next 20 frames as network predictions. During testing, the last 30 frames of the first half of each video are used as condition frames, and the model autoregressively predicts the second half. Metrics for each scene are averaged over sequences from the 10 seen viewpoints.
Note that, ExtDM is a video generation model and cannot evaluate novel-view future frame prediction.

For MAGI-1\cite{AI2025}, we use the official pretrained MAGI-1-4.5B-distill model with 16 sampling steps and 32 condition frames. For each test video, the last 32 frames of the first half serve as condition frames, and the model predicts the second half directly. For all 103 test scenes, metrics are averaged across the 10 seen viewpoints.

\subsubsection{Details of Continuous Short-term Prediction}
We randomly select 103 scenes from validation set. Each scene includes 10 seen viewpoints and 2 novel viewpoints. For every scene, we evaluate the model’s future frame prediction performance on both seen and novel viewpoints using four metrics: PMF, PSNR, SSIM, LPIPS. Starting from $t=0$, at each time step, we incorporate new observations to update the model (or treat them as additional condition frames), then predict the next 10 frames for both seen and novel viewpoints. Metrics are computed for each prediction and averaged across seen views and novel views. If a video contains $N$ frames, we aggregate results over $t=0,\dots,N-10$ and report the mean score for seen and novel viewpoints.

To enable continuous prediction, we modify the training process for FreeGave and DefGS. At each incremental step, the model observes new frames from seen viewpoints, updates its parameters, predicts the next 10 frames, and computes averaged metrics for seen and novel viewpoints. Final scores are obtained by averaging metrics across all incremental steps. For ExtDM and MAGI-1, new observations are treated as additional conditioning frames.

FreeGave: For each scene, we use 10 viewpoints as training views. Initialization follows the Long-term Prediction setup: the first frame is optimized for 3000 iterations using vanilla Gaussian Splatting to reconstruct a static scene. For each new time step, FreeGave updates its velocity field prediction using newly observed frames. 

DefGS: Initialization follows the settings in Long-term Prediction: the first frame undergoes 3000 iterations of optimizing vanilla Gaussians. For each new observation, we optimize the deformation field for 50 iterations and Gaussian kernels for 150 iterations. At each incremental step, we predict 10 future frames and compute metrics as the average over 10 seen viewpoints 
and 2 novel viewpoints.

ExtDM: We directly evaluate the model trained for Long-term Prediction. The number of condition frames is set as 30; if fewer frames are available, the first frame is repeated to reach 30. Only seen viewpoints are evaluated.

MAGI-1: We adopt the same settings used in Long-term Prediction. The number of condition frames is set as 32; if insufficient, the first frame is repeated to reach 32. Only seen viewpoints are evaluated.

\subsection{More Details of Physical Properties Estimation}\label{app:phys_esti_exp}
In this experiment, we evaluate two representative baselines, PAC-NeRF~\cite{Li2023} and GIC~\cite{Cai2024}, on our dataset.  
Both methods follow a reconstruction-then-regression strategy: first reconstructing the dynamic 3D scene from multi-view video observations, and then regressing the relevant physical parameters from the temporal evolution of the reconstructed representation, according to those predefined evolution equations in Section~\ref{sec:taichi_lang}.

Specifically, PAC-NeRF learns a neural radiance field comprising a view-independent density field and color field. It then estimates the initial velocity and physical parameters (\eg{}, viscosity, Young’s modulus) via a regression module that analyzes the temporal dynamics of the learned radiance fields.  
Similarly, GIC reconstructs the full dynamic 3D scene using dynamic 3D Gaussian network and subsequently regresses the physical parameters from the optimized dynamic Gaussian representation.

To evaluate these methods, we select 20 scenes from our test set, ensuring uniform representation across all five material categories (\ie{}, 4 samples per material). For reconstruction, we use 13 out of the 15 available camera views as training views, reserving the remaining 2 views as novel views for evaluation.  
For the two baselines, we provide 90 frames for five materials. We evaluate the two methods on the whole frames in novel-view rendering and resimulation. Since previous datasets for physical properties estimation leverage approximately 20 frames to simulate the whole moving progress, we alternatively downsample the frames to seek the best results for PAC-NeRF and GIC during the training process. In addition, our dataset provides diverse cases with complex boundaries. We also include the corresponding SDF for the boundaries during training and testing. As shown in Table \ref{tab:phys_prop_init}, we provide the initial guess for five types of materials, which is identical to the original settings from the two methods.

\begin{table}[htb] \tabcolsep=0.12cm \vspace{-0.05cm}
\centering
\caption{The initial guess of physical parameters estimation for five types of materials on \nickname{}. }\vspace{-0.3cm}
\label{tab:phys_prop_init}
\resizebox{0.5\textwidth}{!}{
\begin{tabular}{l|ccccl}
\rowcolor{headergray}
\hline
                                        & Elastic Solids                                                 & Plasticine                                                                     & Newtonian Fluids                                               & Non-Newtonian Fluids                                                                      & \multicolumn{1}{c}{Granular Substances} \\ \hline
PAC-NeRF \cite{Li2023} & \begin{tabular}[c]{@{}c@{}} $E = 316228$ \\ $\nu = 0.25$ \end{tabular} & \begin{tabular}[c]{@{}c@{}} $E  = 10000$ \\ $\nu = 0.25$ \\ $\tau_{y} = 1000$ \end{tabular} & \begin{tabular}[c]{@{}c@{}} $\mu = 10$ \\ $\kappa = 1000$ \end{tabular} & \begin{tabular}[c]{@{}c@{}} $ \mu = 100$ \\ $\kappa = 100000$ \\ $\tau_{y} = 10$ \\ $\eta = 1$ \end{tabular} & $\theta_{fric} = 10$                       \\ \hline
GIC \cite{Cai2024}  & \begin{tabular}[c]{@{}c@{}} $E = 316228$ \\ $\nu = 0.25$ \end{tabular} & \begin{tabular}[c]{@{}c@{}} $E  = 10000$ \\ $\nu = 0.25$ \\ $\tau_{y} = 1000$ \end{tabular} & \begin{tabular}[c]{@{}c@{}} $\mu = 10$ \\ $\kappa = 1000$ \end{tabular} & \begin{tabular}[c]{@{}c@{}} $ \mu = 100$ \\ $\kappa = 100000$ \\ $\tau_{y} = 10$ \\ $\eta = 1$ \end{tabular} & $\theta_{fric} = 10$                       \\ \hline
\end{tabular}
}\vspace{-0.3cm}
\end{table}

To visualize the results, we render the reconstructed scenes from novel views and resimulate the dynamics using the estimated physical parameters.  
In resimulation, we slightly perturb the initial position or orientation to demonstrate the model’s ability to generalize beyond the original configuration. 

We report additional quantitative results for novel-view rendering and resimulation in Table~\ref{tab:phys_prop_esti_novelview_app} and Table~\ref{tab:phys_prop_esti_resim_app}, respectively, with performance broken down by material types.

\begin{table}[htb] \tabcolsep=0.2cm  \vspace{-0.3cm}
\centering
\caption{Additional quantitative results of novel view synthesis based on estimated physical properties.}\vspace{-0.25cm}
\label{tab:phys_prop_esti_novelview_app}
\resizebox{0.48\textwidth}{!}{
\begin{tabular}{ll|cccc}
\toprule
\rowcolor{headergray} 
 & & PMF $\uparrow$ & PSNR $\uparrow$ & SSIM $\uparrow$ & LPIPS$\downarrow$ \\ \hline
\multirow{2}{*}{Elastic Solids} & PAC-NeRF \cite{Li2023} & 5.158 & 27.909 & 0.966 & 0.062 \\
& GIC \cite{Cai2024}  & \textbf{5.570} & \textbf{30.143} & \textbf{0.968} & \textbf{0.057} \\ \hline

\multirow{2}{*}{Plasticine} & PAC-NeRF \cite{Li2023} & 5.071 & 27.429 & 0.967 & 0.051 \\
& GIC \cite{Cai2024}  & \textbf{5.535} & \textbf{30.191} & \textbf{0.971} & \textbf{0.044} \\ \hline

\multirow{2}{*}{Newtonian Fluids} & PAC-NeRF \cite{Li2023} & 4.201 & 24.797 & 0.957 & 0.061 \\
& GIC \cite{Cai2024}  & \textbf{5.047} & \textbf{29.381} & \textbf{0.967} & \textbf{0.050} \\ \hline

\multirow{2}{*}{Non-Newtonian Fluids} & PAC-NeRF \cite{Li2023} & 4.197 & 24.696 & 0.959 & 0.064 \\
& GIC \cite{Cai2024}  & \textbf{4.694} & \textbf{27.617} & \textbf{0.969} & \textbf{0.056} \\ \hline

\multirow{2}{*}{Granular Substances} & PAC-NeRF \cite{Li2023} & \textbf{4.289} & 24.098 & 0.859 & 0.191 \\
& GIC \cite{Cai2024}  & 4.185 & \textbf{25.033} & \textbf{0.863} & \textbf{0.180} \\

\bottomrule
\end{tabular}
}\vspace{-0.2cm}
\end{table}

\begin{table}[htb] \tabcolsep=0.2cm  \vspace{-0.3cm}
\centering
\caption{Additional quantitative results of resimulation based on estimated physical properties.}\vspace{-0.25cm}
\label{tab:phys_prop_esti_resim_app}
\resizebox{0.48\textwidth}{!}{
\begin{tabular}{ll|cccc}
\toprule
\rowcolor{headergray} 
 & & PMF $\uparrow$ & PSNR $\uparrow$ & SSIM $\uparrow$ & LPIPS$\downarrow$ \\ \hline
\multirow{2}{*}{Elastic Solids} & PAC-NeRF \cite{Li2023} & 4.701 & 25.463 & 0.964 & 0.062 \\
& GIC \cite{Cai2024}  & \textbf{5.136} & \textbf{27.886} & \textbf{0.967} & \textbf{0.055} \\ \hline

\multirow{2}{*}{Plasticine} & PAC-NeRF \cite{Li2023} & 4.614 & 24.850 & 0.963 & 0.058 \\
& GIC \cite{Cai2024}  & \textbf{5.206} & \textbf{28.332} & \textbf{0.969} & \textbf{0.047} \\ \hline

\multirow{2}{*}{Newtonian Fluids} & PAC-NeRF \cite{Li2023} & 4.166 & 24.253 & 0.956 & 0.061 \\
& GIC \cite{Cai2024}  & \textbf{4.900} & \textbf{28.144} & \textbf{0.964} & \textbf{0.051} \\ \hline

\multirow{2}{*}{Non-Newtonian Fluids} & PAC-NeRF \cite{Li2023} & 3.999 & 22.415 & 0.946 & 0.079 \\
& GIC \cite{Cai2024}  & \textbf{4.543} & \textbf{25.675} & \textbf{0.966} & \textbf{0.058} \\ \hline

\multirow{2}{*}{Granular Substances} & PAC-NeRF \cite{Li2023} & 4.174 & 23.615 & 0.882 & 0.167 \\
& GIC \cite{Cai2024}  & \textbf{4.202} & \textbf{24.477} & \textbf{0.884} & \textbf{0.157} \\
\bottomrule
\end{tabular}
}\vspace{-0.4cm}
\end{table}

\subsection{More Details of Motion Transfer}\label{app:motion_transfer_exp}
Motion transfer refers to the task of transferring motion dynamics from a reference video to a static image, generating a video that inherits the motion pattern of the reference while preserving the appearance and semantic identity of the object(s) in the given image.

Current approaches typically represent motion through cross-frame pixel correspondences, \ie{}, optical flow, which encode per-pixel displacement fields across time. For instance, MotionPro\cite{Zhang2025} employs an optical flow analyzer to extract motion trajectories from the reference video and injects these motion features into an image-to-video generation pipeline. Similarly, GoWithTheFlow\cite{Burgert2025} introduces a novel noise-warping mechanism that incorporates optical flow directly into the noise initialization stage of the diffusion process, thereby steering the denoising trajectory to replicate the reference motion in the generated output.

In our experiments, we select 273 motion-transfer pairs from our validation set, where the only difference between the reference video and the target video lies in the primary moving object (\eg{}, different instances of a bouncing ball or falling toy). The condition image is the first frame of the target video. The text prompt, if used, is the one originally paired with the target video in our dataset.